\documentclass[fleqn,10pt]{wlscirep}
\usepackage[utf8]{inputenc}
\usepackage[T1]{fontenc}
\usepackage{subcaption}
\usepackage{multirow}
\usepackage{threeparttable}
\usepackage{array}
\usepackage{supertabular}
\usepackage{algorithm}
\usepackage{algorithmicx}
\usepackage{algpseudocode}
\usepackage{caption}
\usepackage{hyperref}
\usepackage[cal=cm]{mathalfa}

\title{Bridging Data Gaps of Rare Conditions in ICU: A Multi-Disease Adaptation Approach for Clinical Prediction}

\author[1,*]{Mingcheng Zhu}
\author[1]{Yu Liu}
\author[1]{Zhiyao Luo}
\author[1]{Tingting Zhu}
\affil[1]{Department of Engineering Science, University of Oxford, Oxford, OX1 2JD, UK}

\affil[*]{mingcheng.zhu@eng.ox.ac.uk}

\begin{abstract}
Artificial Intelligence has revolutionised critical care for common conditions. Yet, rare conditions in the intensive care unit (ICU), including recognised rare diseases and low-prevalence conditions in the ICU, remain underserved due to data scarcity and intra-condition heterogeneity. To bridge such gaps, we developed KnowRare, a domain adaptation-based deep learning framework for predicting clinical outcomes for rare conditions in the ICU. KnowRare mitigates data scarcity by initially learning condition-agnostic representations from diverse electronic health records through self-supervised pre-training. It addresses intra-condition heterogeneity by selectively adapting knowledge from clinically similar conditions with a developed condition knowledge graph. Evaluated on two ICU datasets across five clinical prediction tasks (90-day mortality, 30-day readmission, ICU mortality, remaining length of stay, and phenotyping), KnowRare consistently outperformed existing state-of-the-art models. Additionally, KnowRare demonstrated superior predictive performance compared to established ICU scoring systems, including APACHE IV and IV-a. Case studies further demonstrated KnowRare's flexibility in adapting its parameters to accommodate dataset-specific and task-specific characteristics, its generalisation to common conditions under limited data scenarios, and its rationality in selecting source conditions. These findings highlight KnowRare's potential as a robust and practical solution for supporting clinical decision-making and improving care for rare conditions in the ICU.
\end{abstract}
\begin{document}

\flushbottom
\maketitle
%
%
\thispagestyle{empty}

\section*{Introduction}
Rare conditions in the intensive care unit (ICU), including both formally classified rare diseases and low-prevalence conditions in the ICU \cite{wang2024operational}, can be life-threatening or chronically debilitating, contributing to a high burden on health systems \cite{mao2025phenotype}. Patients with rare conditions often face challenges such as limited access to specialised clinical expertise, frequent misdiagnoses, and prolonged diagnostic delays \cite{willmen2023rare, benito2022diagnostic}. These factors collectively contribute to worse clinical outcomes, such as a longer stay in the ICU, higher readmission rates, and increased post-discharge mortality compared to common conditions \cite{blazsik2021impact, mazzucato2023estimating}. Consequently, rare conditions place a greater per-patient strain on critical care resources than common conditions.

Although artificial intelligence, especially deep learning (DL), has significantly advanced critical care analytics for common conditions such as septic shock and heart failure \cite{wang2025artificial, zhang2024ai}, its application to rare conditions in the ICU remains limited. Existing DL models for common conditions frequently underperform for rare conditions due to insufficient training data, preventing the development of robust and generalisable predictive models \cite{zhao2024leave}. Adding to this problem, the geographic dispersion of patients contributes to the variability in clinical practices and observations across institutions \cite{blazsik2021impact}. Meanwhile, rare conditions often exhibit complex clinical manifestations due to multisystem involvement, affecting multiple organs or physiological pathways \cite{mao2025phenotype}. Collectively, these factors result in substantial intra-condition heterogeneity. 

Recent efforts to improve predictive performance for rare conditions focused mainly on overcoming data scarcity, employing approaches such as few-shot learning \cite{zhang2019metapred, tan2022metacare++}, federated learning~\cite{chen2023dfml, meduri2025leveraging}, large-scale pre-training \cite{zhao2024leave, prakash2021rarebert, yu2024smart, west2025unsupervised, zhao2025unveiling}, and synthetic data generation \cite{ma2020rare, li2023mlgan}. Although these approaches have shown potential in mitigating data scarcity, they often overlook intra-condition heterogeneity. Moreover, existing methods develop one-size-fits-all models, which can work for all conditions or a mixture of conditions. This paradigm further compromises model performance, especially given the diverse presentations and pathophysiology of rare conditions \cite{oakden2020hidden}. Therefore, a model designed to bridge the gaps caused by data scarcity and intra-condition heterogeneity in the prediction of outcomes for rare conditions is critical.

In this study, we introduce KnowRare, a DL framework specifically designed to bridge the challenges posed by data scarcity and intra-condition heterogeneity in rare conditions. To address data scarcity, KnowRare first learns condition-agnostic representations through self-supervised pretraining across diverse conditions. To address intra-condition heterogeneity, KnowRare employs knowledge-guided domain adaptation, enabling it to learn robust representations that capture the heterogeneity within the rare conditions. This adaptation is guided by a condition knowledge graph (KG), which encodes clinical similarities among conditions and enables the selective transfer of relevant knowledge from clinically similar conditions. The framework comprises three key modules: knowledge-guided domain selection to identify clinically similar source conditions, condition-agnostic pretraining to establish robust time-series representations, and joint adversarial domain adaptation to align variable and outcome distributions across selected conditions. These modules enable KnowRare to provide robust condition-specific predictions for rare conditions in the ICU.

We validated KnowRare for five clinical prediction tasks, including 90-day mortality, 30-day readmission, ICU mortality, remaining length of stay (LoS), and phenotyping, using the MIMIC-III and eICU datasets~\cite{johnson2016mimic, pollard2018eicu}. KnowRare consistently outperformed baseline methods in predicting outcomes for rare conditions, demonstrating superior predictive performance and robustness. These findings highlight the effectiveness of KnowRare in overcoming critical gaps caused by data scarcity and intra-condition heterogeneity. By providing accurate, disease-specific predictions, KnowRare has implications for clinical decision-making, improving patient outcomes, and optimising resource allocation tailored to rare conditions in the ICU.

\section*{Results}

\begin{figure}
  \includegraphics[width=0.98\textwidth]{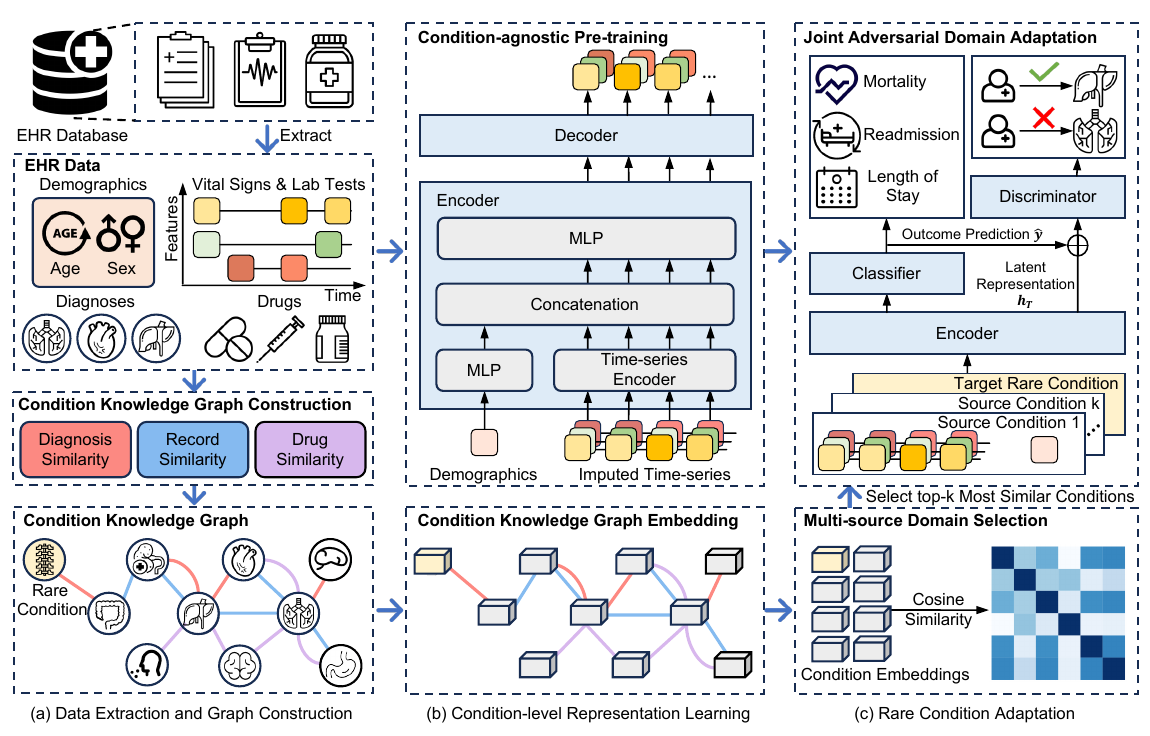}
  \vspace{-6px}
  \caption{\textbf{Overview of the KnowRare framework.} KnowRare operates in three steps: \textbf{(a)} \textit{Data extraction and graph construction}: Structured EHR data, including demographic data, vital signs, laboratory tests, diagnoses, and drug records, are extracted and preprocessed through aggregation, imputation, and normalisation. Condition similarities are quantified from three perspectives: diagnosis co-occurrence (categorical), record variable distributions (continuous), and shared drug usage (categorical). These similarities are integrated into a heterogeneous condition knowledge graph (KG), capturing comprehensive clinical relationships. \textbf{(b)} \textit{Condition-level representation learning}: This step involves two modules. The condition KG embedding module uses KG embedding techniques to generate condition embeddings that represent clinical similarities among conditions. Concurrently, the condition-agnostic pre-training module trains a time-series encoder via self-supervision to learn general temporal patterns independent of specific conditions, providing robust initial latent representations. \textbf{(c)} \textit{Rare condition adaptation}: This step includes two modules. First, the knowledge-guided domain selection module identifies the top-$k$ source conditions most similar to the target rare condition by calculating cosine similarity between condition embeddings. Subsequently, the joint adversarial domain adaptation module fine-tunes the pre-trained time-series encoder with the target rare condition and the selected top-$k$ source conditions. The encoder produces patient-level latent representations ($h_T$), integrating condition-agnostic knowledge with insights derived from similar source conditions. Based on these latent representations, a classifier predicts clinical outcomes ($\hat{y}$; mortality, readmission, length-of-stay, etc.). Concurrently, a discriminator network is trained adversarially to distinguish whether the latent representations and predicted outcomes originate from patients with the target rare condition or the selected source conditions. This adversarial process ensures the encoder generates robust representations, which improve predictive performance for heterogeneous rare conditions in the ICU.}
\label{fig:method}
\end{figure}

\subsection*{Overview of KnowRare}
 KnowRare (Figure~\ref{fig:method}) addresses the challenges of scarcity and heterogeneity in predicting the outcomes of rare conditions by integrating clinical knowledge across conditions. Initially, structured patient data are extracted from EHR databases and systematically preprocessed. Condition similarities are quantified from three complementary clinical perspectives: diagnosis co-occurrence, statistical similarity of patient records, and shared medication usage. These similarity measures are combined into a heterogeneous condition KG, providing a structured representation of clinical relationships among conditions. Subsequently, KnowRare performs condition-level representation learning through two complementary strategies. First, condition embeddings are generated with a KG embedding method on the condition KG, capturing meaningful clinical relationships between conditions. Second, a time-series encoder is pre-trained with the self-supervised next-step prediction, learning generalisable temporal patterns from patient time-series data across all conditions. Finally, KnowRare adapts its learnt representations to individual rare conditions by selecting the top-$k$ most similar conditions based on the cosine similarity of the condition embeddings. Joint adversarial domain adaptation is then employed, aligning both latent representations and prediction outcomes between the selected conditions and the target rare condition. This targeted adaptation enables condition-specific predictions with transferable knowledge from similar conditions.

\subsection*{Data description}
We evaluated KnowRare using two publicly available ICU EHR datasets: MIMIC-III \cite{johnson2016mimic} and eICU \cite{pollard2018eicu}, both providing granular clinical records suitable for studying rare conditions in the ICU. MIMIC-III is a large, single-centre critical care database comprising health-related data from 46,520 patients admitted to Beth Israel Deaconess Medical Centre between 2001 and 2012. In contrast, eICU is a multi-centre dataset collected from more than 200 hospitals, offering broader variability in patient demographics, phenotypes, and treatment protocols. Following established conventions \cite{nguengang2020estimating, liu2022natural}, we classified conditions as \textit{Rare} if their prevalence is fewer than one case per 2,000 patients within each dataset. The statistics of the datasets are summarised in Table~\ref{tab:dataset_statistics}, and the variables extracted from each dataset are detailed in Supplementary Table~\ref{table:variables}.

\begin{table}[h!]
\centering
\caption{Statistics for EHR Databases}
\label{tab:dataset_statistics}
\begin{tabular}{l |c c}
\hline
\textbf{Statistic} & MIMIC-III & eICU \\ \hline
Number of patients & 46,520 & 139,367 \\ 
Number of hospital Visits & 58,976 & 166,355 \\
Number of ICU stays & 61,532 & 200,859 \\ 
Number of hospitals & 1 (single-centre) & 208 (multi-centre)\\
Number of level 3 ICD-9-CM codes & 587 & 303 \\ 
Number of Rare Conditions (Percentage) & 383 (65.2\%) & 192 (63.4\%)\\
\hline
\end{tabular}
\end{table}

For MIMIC-III, we evaluated two prediction tasks: (1) 30-day readmission, predicting whether a patient will be readmitted to the hospital within 30 days after discharge, using the final 48 hours of hospital admission data; and (2) 90-day mortality, predicting whether a patient will die within 90 days after discharge, also using the final 48 hours of hospital admission data. For eICU, we assessed three tasks: (1) ICU mortality, predicting whether a patient will die during their ICU stay, using the first 24 hours of ICU admission data; (2) length of stay (LoS), predicting the remaining duration of stay in the ICU using the initial 24 hours of data, with targets categorised into 10 intervals: less than 1 day, 1–2 days, 2–3 days, 3–4 days, 4–5 days, 5–6 days, 6–7 days, 7–10 days, 10–14 days, and more than 14 days; and (3) classification of phenotypes, predicting the presence of 25 acute care phenotypes using the first 24 hours of ICU admission data. All tasks adhere to established benchmark definitions \cite{gupta2022extensive, harutyunyan2019multitask}. 

We selected the ten least prevalent conditions in the ICU, including both recognised rare diseases and low-prevalence conditions,  based on two criteria: (1) prevalence of the condition less than one in 2,000 patients \cite{nguengang2020estimating, schaefer2020use}, and (2) at least one positive outcome sample (e.g., mortality or readmission case) per prediction task evaluated. The conditions were ranked according to their total number of cases, and the least prevalent ones satisfying both criteria were selected until we obtained ten qualifying conditions (the statistics are summarised in Supplementary Tables~\ref{tab:mimic_rare_stats} and \ref{tab:eicu_rare_stats}). To prevent data leakage, patients were split into training (67\%), validation (16\%), and test (17\%) sets, maintaining patient-level separation across sets \cite{yang2021safedrug, zhao2024leave}. Additional details on preprocessing steps are provided in Supplementary~\ref{appendix sec:data processing}. Model performance for these rare conditions was evaluated using the Area Under the Receiver Operating Characteristic Curve (AUROC) and the Area Under the Precision-Recall Curve (AUPRC), with the latter preferred for class imbalance.

\subsection*{Performance evaluation of KnowRare}

\begin{table}[ht!]
\centering
\fontsize{6}{7.5}\selectfont
\begin{threeparttable}
\caption{Performance comparison of KnowRare with other baselines. }
\hspace*{-0.4cm}
\label{table:comparison}
    \begin{tabular}{l|cc | cc| cc | cc | cc}
    \hline
    \multirow{3}{*}{\textbf{Method}} & \multicolumn{4}{c|}{\textbf{MIMIC-III}} & \multicolumn{6}{c}{\textbf{eICU}} \\ 
    \cline{2-5} \cline {6-11}
    &\multicolumn{2}{c|}{\textbf{90 Days Mortality}} & \multicolumn{2}{c|}{\textbf{30 Days Readmission}} & \multicolumn{2}{c|}{\textbf{ICU Mortality}} & \multicolumn{2}{c|}{\textbf{Remaining LoS}} & \multicolumn{2}{c}{\textbf{Phenotyping}}\\
    &\textbf{AUPRC} & \textbf{AUROC} & \textbf{AUPRC} & \textbf{AUROC}& \textbf{AUPRC} & \textbf{AUROC}  & \textbf{AUPRC} & \textbf{AUROC} & \textbf{AUPRC} & \textbf{AUROC} \\
    \hline
    LSTM\cite{hochreiter1997long} (single) &0.339 (0.039) & 0.432 (0.072) & 0.370 (0.044) & 0.488 (0.062) & 0.427 (0.113) & 0.491 (0.106) & 0.125 (0.022) & 0.459 (0.061) & 0.131 (0.009) & 0.510 (0.089) \\
    LSTM\cite{hochreiter1997long} (all) & 0.725 (0.028) & 0.789 (0.022) & 0.693 (0.095) & 0.744 (0.048) & 0.569 (0.039) & 0.725 (0.051) & 0.124 (0.058)& \underline{0.519} (0.037) & 0.137 (0.032) & 0.555 (0.040)\\
    Transformer\cite{vaswani2017attention} (single) & 0.539 (0.130) & 0.610 (0.126) & 0.439 (0.085) & 0.578 (0.073) & 0.381 (0.058) & 0.521 (0.055) & 0.135 (0.017) & 0.491 (0.056) & 0.152 (0.058) & 0.476 (0.074)\\
    Transformer\cite{vaswani2017attention} (all) & 0.665 (0.045) & 0.751 (0.048) & 0.706 (0.031) & \underline{0.747} (0.014) & \underline{0.636} (0.013) & \textbf{0.770} (0.023) & 0.090 (0.066)& 0.505 (0.009) &0.121 (0.033) & 0.597 (0.056)\\
    RETAIN \cite{choi2016retain} (single) & 0.365 (0.027) & 0.493 (0.056) & 0.311 (0.019) & 0.448 (0.055) & 0.415 (0.074) & 0.518 (0.042) & 0.141 (0.027) & 0.514 (0.053) & 0.172 (0.022) & 0.545 (0.032) \\
    RETAIN \cite{choi2016retain} (all) & 0.679 (0.058) & 0.770 (0.032) & 0.545 (0.082) & 0.625 (0.086) & 0.532 (0.046) & 0.733 (0.025) & NA & NA & 0.061 (0.041) & 0.488 (0.032)\\
    \hline
    Metapred \cite{zhang2019metapred} & 0.481 (0.028)& 0.618 (0.031) & 0.394 (0.044)& 0.534 (0.016) & 0.557 (0.038) & 0.643 (0.024) & \underline{0.176} (0.097) & \textbf{0.525} (0.025) & \underline{0.237} (0.062) & \textbf{0.656} (0.026)\\
    RareMed \cite{zhao2024leave} & 0.565 (0.028) & 0.773 (0.023) & 0.686 (0.050) & 0.741 (0.052) & 0.545 (0.018)& 0.625 (0.032) & 0.096 (0.039) & 0.494 (0.008) & 0.168 (0.017) & 0.591 (0.039) \\
    SMART \cite{yu2024smart} & 0.610 (0.062) & 0.733 (0.035) & 0.705 (0.056) & 0.721 (0.039) & 0.544 (0.027) & 0.711 (0.025) & 0.059 (0.021) & 0.496 (0.019) &0.220 (0.035) & 0.565 (0.044)\\
    FADA \cite{teshima2020few} & \underline{0.737} (0.024) & 0.784 (0.018) & 0.701 (0.044) & 0.730 (0.050) & 0.521 (0.036) & 0.684 (0.030) & 0.057 (0.001) & 0.500 (0.0) & 0.156 (0.032) & 0.540 (0.027) \\
    Advdiag \cite{zhang2022adadiag} & 0.717 (0.023) & 0.779 (0.004) & \underline{0.711} (0.075) & 0.730 (0.045) & 0.497 (0.026) & 0.631 (0.033) & 0.094 (0.044) & 0.494 (0.030) & 0.157 (0.024) & 0.532 (0.027)\\
    Stable-CRP \cite{lee2023stable} & 0.726 (0.041)& \textbf{0.819} (0.027) & 0.646 (0.069)& 0.686 (0.062) & 0.631 (0.053)& 0.691 (0.027) & 0.077(0.027) & 0.501 (0.026) & 0.158 (0.034) & 0.546 (0.029)  \\
    MANYDG \cite{yang2023manydg} & 0.622 (0.054) & 0.663 (0.079) & 0.637 (0.068) & 0.643 (0.038) & 0.470 (0.054) & 0.632 (0.036) & 0.138 (0.009) & 0.453 (0.014) & 0.210 (0.056) & 0.570 (0.010) \\
    
    \noalign{\vskip 1.5pt}
    \hline
    \textit{KnowRare} (Ours) & \textbf{0.744} (0.051) & \underline{0.797} (0.032) & \textbf{0.716}(0.031) & \textbf{0.749} (0.034) & \textbf{0.709}(0.022) & \underline{0.757} (0.012) & \textbf{0.206} (0.024) & 0.509 (0.030) & \textbf{0.244} (0.039) & \underline{0.609} (0.044) \\
    \noalign{\vskip 1.5pt}
    \hline
    \end{tabular}
\begin{tablenotes}
            \item[1] (single) refers to training the model only on data of rare conditions.
            \item[2] (all) refers to training the model on data of all conditions.
            \item[3] Metrics are reported as mean (std) over five runs with different random seeds. 
            \item[4] The first method group consists of the baselines for ICU outcome predictions, while the second group includes domain adaptation baselines. 
            \item[5] The best results are bolded, and the second-best are underlined.
\end{tablenotes}
\end{threeparttable}
\end{table}

We evaluated the performance of KnowRare across five clinical prediction tasks on two publicly available datasets, comparing it with widely-used methods for ICU outcome prediction and specific methods for limited-data scenarios. As shown in Table~\ref{table:comparison}, KnowRare achieves the highest AUPRC on all five tasks, outperforming both baseline categories. In particular, it surpasses the best domain adaptation method (Stable-CRP) by 12.4\% in ICU mortality (AUPRC: 0.709 vs. 0.631) and outperforms MetaPred by 17.0\% in Remaining LoS (AUPRC: 0.206 vs. 0.176) on the eICU dataset. KnowRare consistently ranks first or second, demonstrating balanced precision-recall trade-offs. 

\subsection*{Performance comparison with ICU scoring systems}

To further evaluate the clinical relevance of KnowRare, we compared its performance in the prediction of ICU mortality with established ICU scoring systems, specifically APACHE IV~\cite{zimmerman2006acute}  and APACHE IV-a~\cite{raschke2018explained}. These systems are considered gold standards for the prediction of mortality in ICU settings~\cite{hu2023interpretable}. The analysis focused on the prediction of ICU mortality in the eICU dataset.

As illustrated in Table~\ref{tab:compare_apache}, KnowRare demonstrated superior performance compared to APACHE IV and APACHE IV-a \cite{zimmerman2006acute}. KnowRare achieved an AUPRC of 0.709 and an AUROC of 0.757, significantly outperforming APACHE IV (AUPRC: 0.639, AUROC: 0.701) and APACHE IV-a (AUPRC: 0.627, AUROC: 0.695). These results indicate that KnowRare provides improved predictive accuracy for ICU mortality, reinforcing its potential clinical utility over traditional scoring systems.

\begin{table}[h!]
    \centering
    \small
    \begin{threeparttable}
    \caption{Performance comparison of KnowRare against ICU scoring systems. }
    \begin{tabular}{l|c c}
    \hline
    \multirow{2}{*}{\textbf{Method}} & \multicolumn{2}{c}{\textbf{ICU Mortality (eICU)}} \\
    &\textbf{AUPRC} & \textbf{AUROC} \\
    \hline
        APACHE IV~\cite{zimmerman2006acute} & 0.639 & 0.701 \\
        APACHE IV-a~\cite{raschke2018explained} & 0.627 & 0.695\\
        \hline
        KnowRare (Ours) & \textbf{0.709} (0.022)& \textbf{0.757} (0.012) \\
    \hline
    \end{tabular}
    \begin{tablenotes}
    \item[1] Metrics are reported as mean (std) over five runs with different random seeds. 
    \end{tablenotes}
    \label{tab:compare_apache}
    \end{threeparttable}
\end{table}

\subsection*{Ablation study}

To assess the contributions of different modules in KnowRare, we conducted an ablation study by independently removing three key modules: \textbf{(1)} knowledge-guided domain selection, \textbf{(2)} condition-agnostic pre-training, and \textbf{(3)} joint adversarial domain adaptation. The results, summarised in Table~\ref{tab:ablation}, indicate that the removal of any module consistently leads to decreased AUROC or AUPRC across all tasks, demonstrating the necessity of each module. Among these, the knowledge-guided domain selection module proved to be the most critical, as its removal resulted in consistent reductions in performance across all tasks on both datasets, with notable performance drops observed in the multi-centre eICU dataset (the mean of AUPRC reduced from 0.709 to 0.573 for ICU mortality and from 0.206 to 0.065 for remaining LoS). Removing condition-agnostic pre-training also consistently reduced performance across most tasks, decreasing the mean of AUPRC for the ICU mortality prediction on MIMIC-III from 0.744 to 0.640. Similarly, removing joint adversarial domain adaptation markedly decreased the mean of AUPRC in the 30-day readmission task on MIMIC-III from 0.716 to 0.481.

\begin{table}[h]
\centering
\fontsize{6}{7.5}\selectfont
\begin{threeparttable}
\caption{Ablation Study of KnowRare.}
\label{tab:ablation}
\hspace*{-0.4cm}
    \begin{tabular}{c| cc | cc | cc | cc | cc }
    \hline
    \multirow{3}{*}{\textbf{Method}} & \multicolumn{4}{c|}{\textbf{MIMIC-III}} & \multicolumn{6}{c}{\textbf{eICU}} \\ 
    \cline{2-5} \cline {6-11}
    &\multicolumn{2}{c|}{\textbf{90 Days Mortality}} & \multicolumn{2}{c|}{\textbf{30 Days Readmission}} & \multicolumn{2}{c|}{\textbf{ICU Mortality}} & \multicolumn{2}{c|}{\textbf{Remaining LoS}} & \multicolumn{2}{c}{\textbf{Phenotyping}} \\
    &\textbf{AUPRC} & \textbf{AUROC} & \textbf{AUPRC} & \textbf{AUROC}& \textbf{AUPRC} & \textbf{AUROC}  & \textbf{AUPRC} & \textbf{AUROC} & \textbf{AUPRC} & \textbf{AUROC} \\
    \hline
    w/o Domain Selection & 0.660 (0.043)& 0.793 (0.015) & 0.699 (0.081) & 0.776 (0.068) & 0.573 (0.026) & 0.661 (0.037) & 0.065 (0.017) & 0.495 (0.011) & 0.146 (0.016) & 0.575 (0.047)\\
    w/o Pre-training & 0.640 (0.090) & 0.730 (0.054)& 0.690 (0.067) & 0.747 (0.031) & 0.628 (0.049) & 0.693 (0.009) & 0.134 (0.011) & 0.483 (0.023) & 0.237 (0.057) & 0.629 (0.032) \\
    w/o Domain Adaptation & 0.679 (0.061) & 0.785 (0.589) & 0.481 (0.139) & 0.598 (0.089) & 0.713 (0.025)& 0.725 (0.013) & 0.151 (0.018) & 0.512 (0.016) & 0.253 (0.049) & 0.586 (0.044) \\
    \hline
    \textit{KnowRare} & 0.744(0.051) & 0.797(0.032)  & 0.716(0.031) & 0.749 (0.034) &0.709(0.022) & 0.757(0.012) & 0.206 (0.024) & 0.509 (0.030) & 0.244 (0.039) & 0.609 (0.044)\\
    \hline
    \end{tabular}
    \begin{tablenotes}
    \item[1] Metrics are reported as mean (std) over five runs with different random seeds. 
    \end{tablenotes}
    \end{threeparttable}
\end{table}

\begin{figure}[htbp]
    \centering

    \begin{minipage}[t]{0.32\textwidth}
        \flushleft
        \textbf{(a)}
    \end{minipage}
    \hfill
    \begin{minipage}[t]{0.32\textwidth}
        \flushleft
        \textbf{(b)}
    \end{minipage}
    \hfill
    \begin{minipage}[t]{0.32\textwidth}
        \flushleft
        \textbf{(c)}
    \end{minipage}
    
    \begin{minipage}[t]{0.32\textwidth}
        \centering
        \setcounter{subfigure}{0}
        \begin{subfigure}[t]{\textwidth}
            \caption*{90d Mortality (MIMIC)}
            \vspace{0.1cm}
            \includegraphics[width=0.95\textwidth]{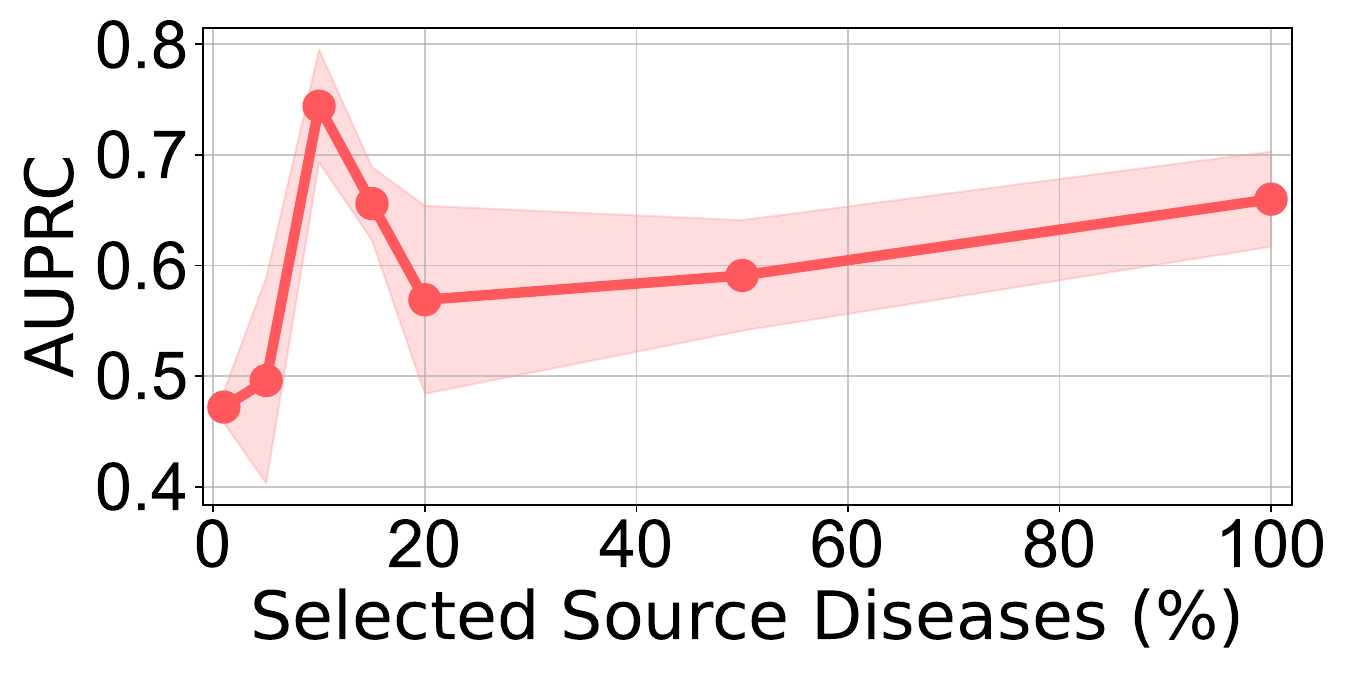}
        \vspace{0.1cm}
        \end{subfigure}
        
        \begin{subfigure}[t]{\textwidth}
            \caption*{30d Readmission (MIMIC)}
            \vspace{0.1cm}
            \includegraphics[width=0.95\textwidth]{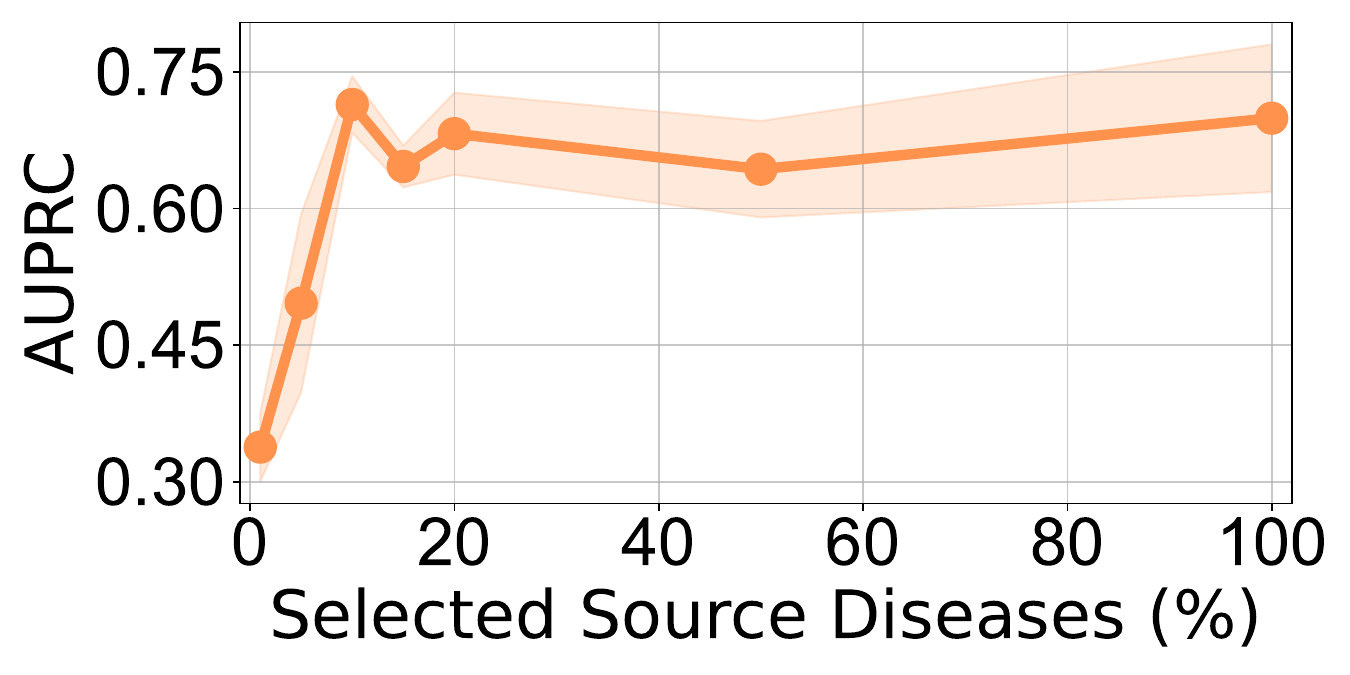}
            \vspace{0.1cm}
        \end{subfigure}
        
        \begin{subfigure}[t]{\textwidth}
        \caption*{ICU Mortality (eICU)}
        \vspace{0.1cm}
        \includegraphics[width=0.95\textwidth]{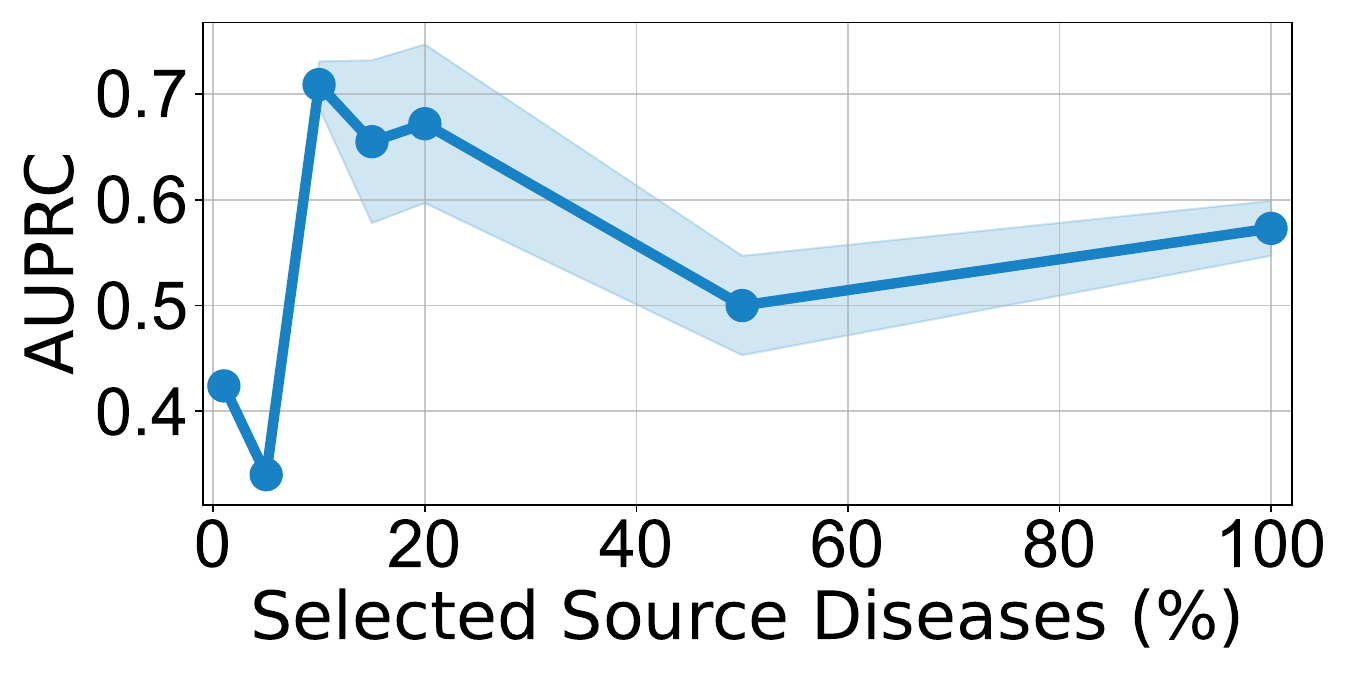}
        \vspace{0.1cm}
        \end{subfigure}

        \begin{subfigure}[t]{\textwidth}
        \caption*{LoS (eICU)}
        \vspace{0.1cm}
        \includegraphics[width=0.95\textwidth]{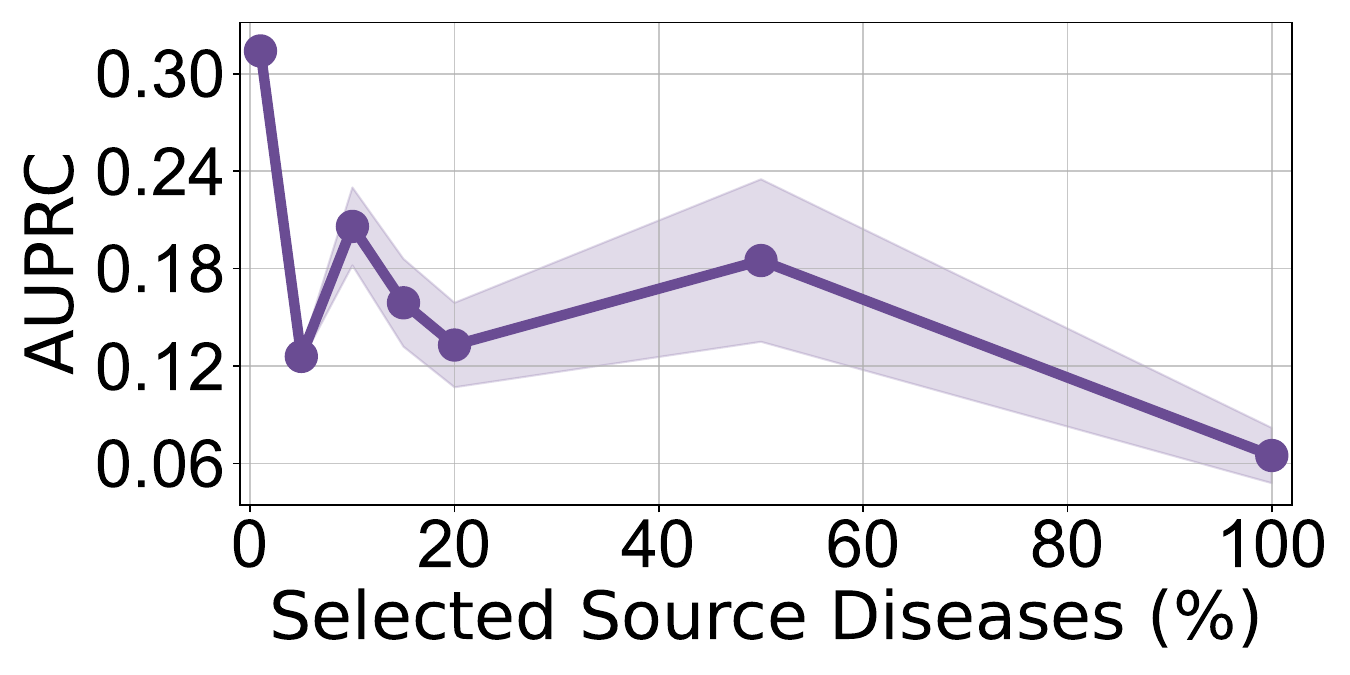}
        \vspace{0.1cm}
        \end{subfigure}

        \begin{subfigure}[t]{\textwidth}
        \caption*{Phenotyping (eICU)}
        \vspace{0.1cm}
        \includegraphics[width=0.95\textwidth]{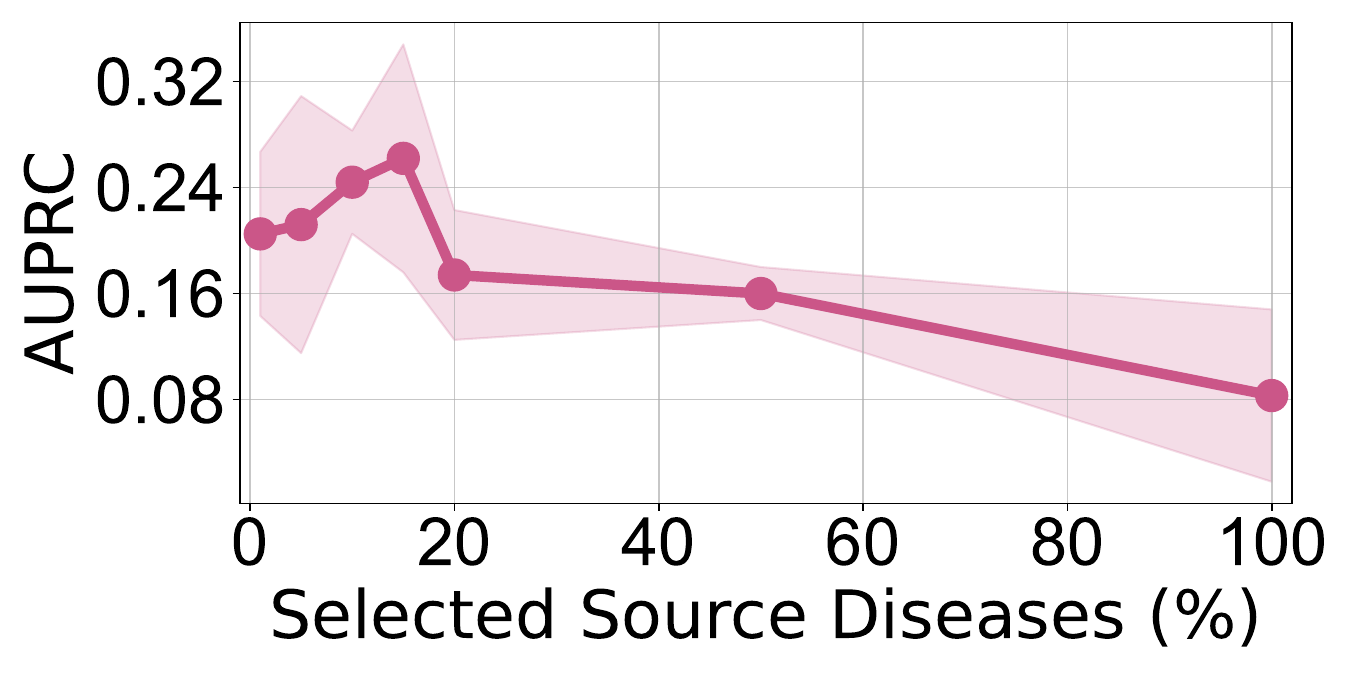}
        \end{subfigure}
    \end{minipage}
    \hfill
    \begin{minipage}[t]{0.32\textwidth}
        \centering
        \setcounter{subfigure}{5}
        \begin{subfigure}[t]{\textwidth}
        \caption*{90d Mortality (MIMIC)}
         \vspace{0.1cm}
            \includegraphics[width=0.95\textwidth]{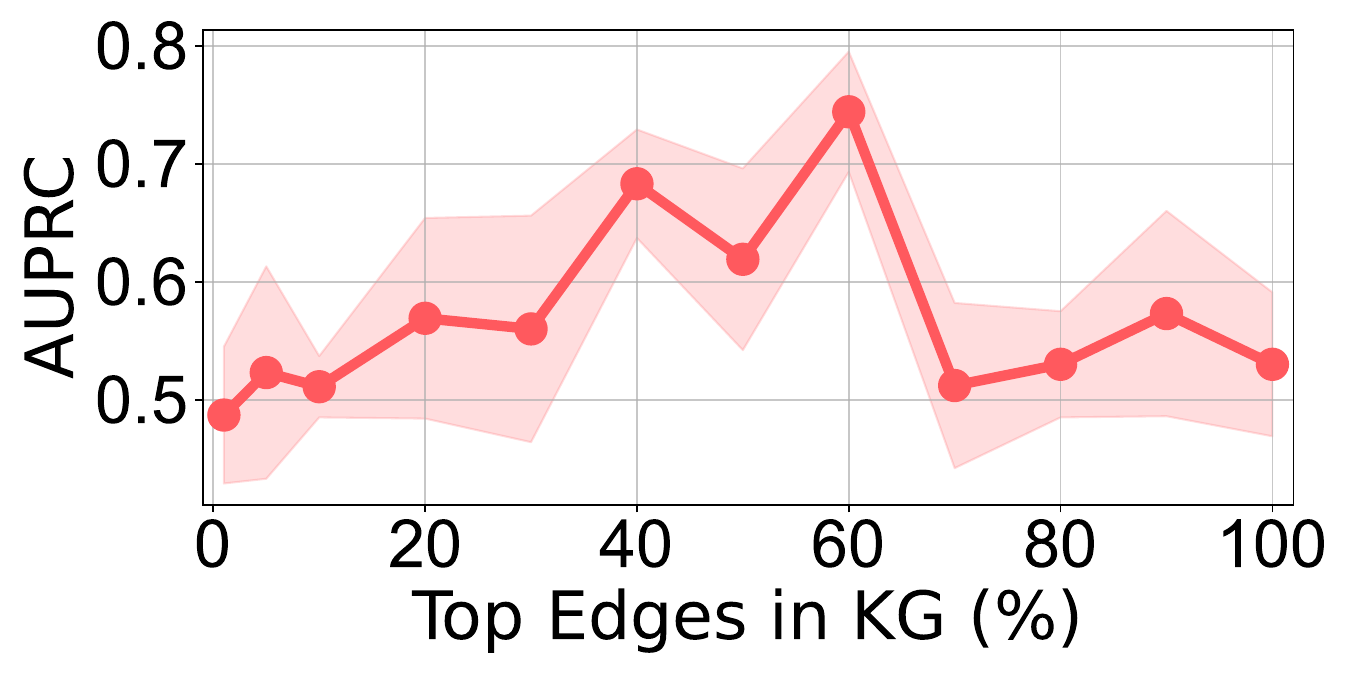}
        \vspace{0.1cm}
        \end{subfigure}
        
        \begin{subfigure}[t]{\textwidth}
            \caption*{30d Readmission (MIMIC)}
            \vspace{0.1cm}
            \includegraphics[width=0.95\textwidth]{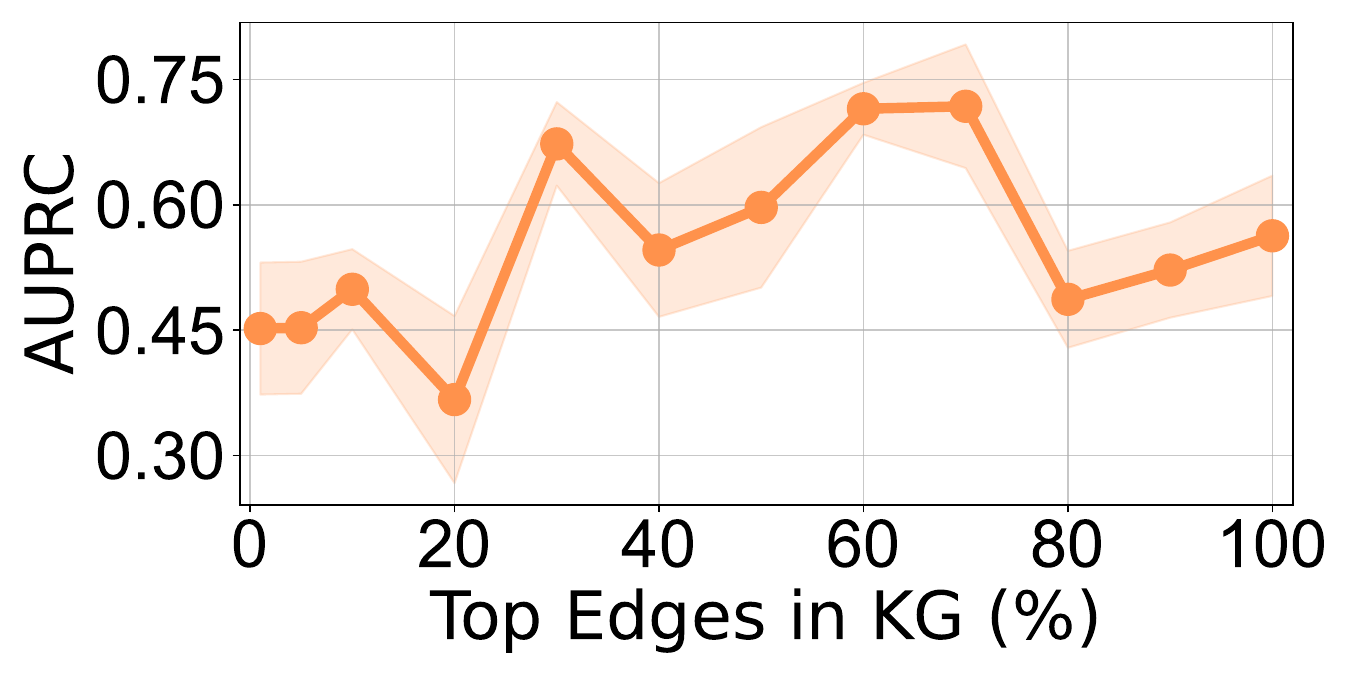}
        \vspace{0.1cm}
        \end{subfigure}
        
        \begin{subfigure}[t]{\textwidth}
        \caption*{ICU Mortality (eICU)}
        \vspace{0.1cm}
            \includegraphics[width=0.95\textwidth]{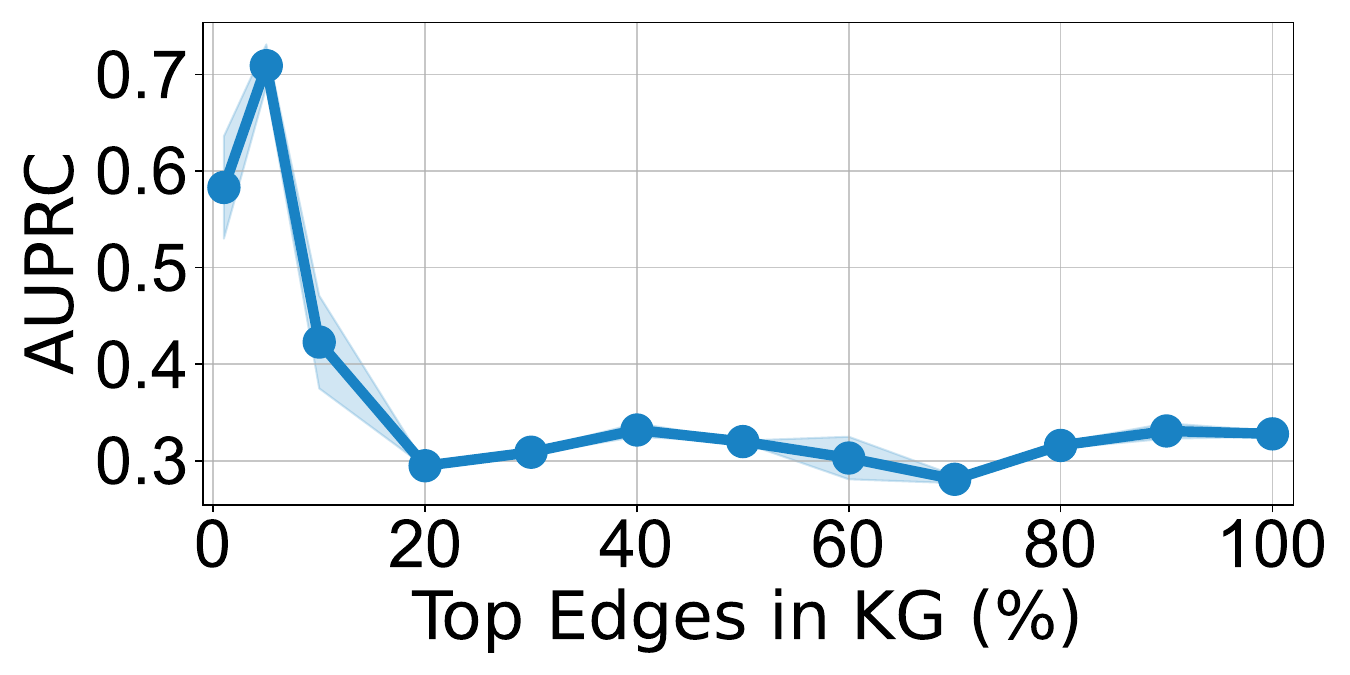}
        \vspace{0.1cm}
        \end{subfigure}
        
        \begin{subfigure}[t]{\textwidth}
        \caption*{LoS (eICU)}
        \vspace{0.1cm}
            \includegraphics[width=0.95\textwidth]{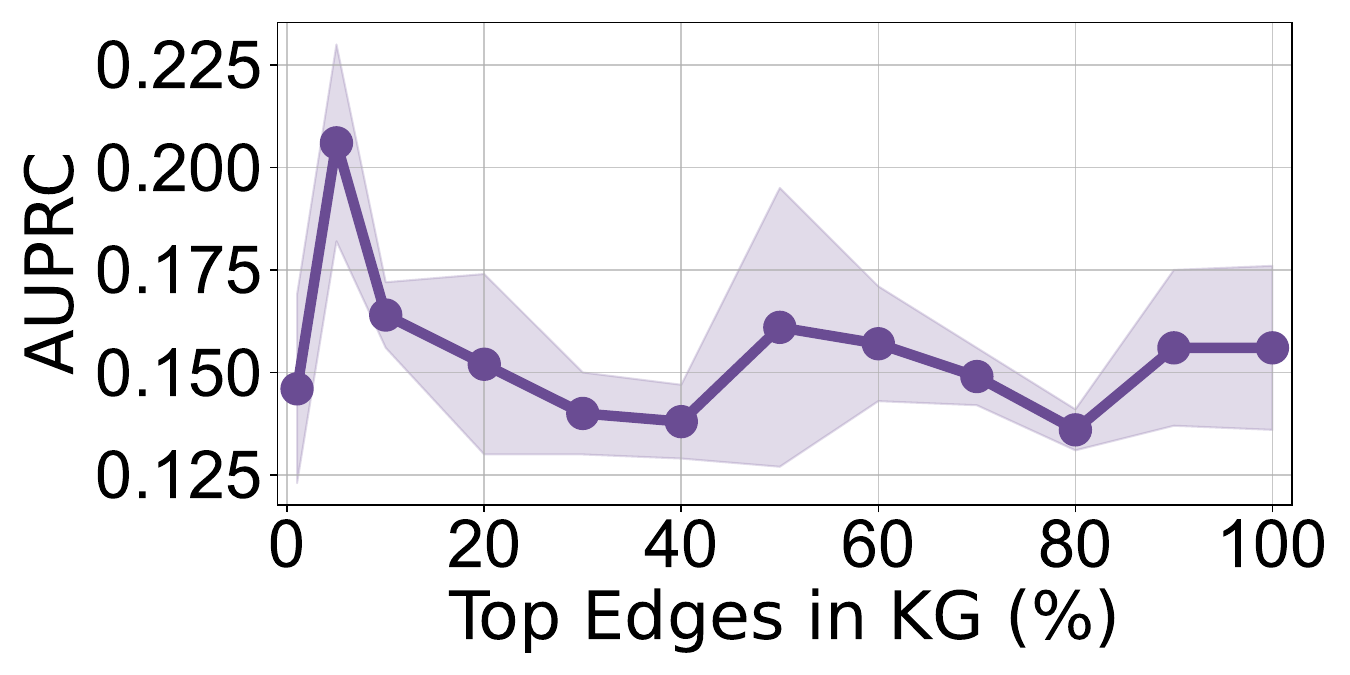}
        \vspace{0.1cm}
        \end{subfigure}
        
        \begin{subfigure}[t]{\textwidth}
        \caption*{Phenotyping (eICU)}
        \vspace{0.1cm}
            \includegraphics[width=0.95\textwidth]{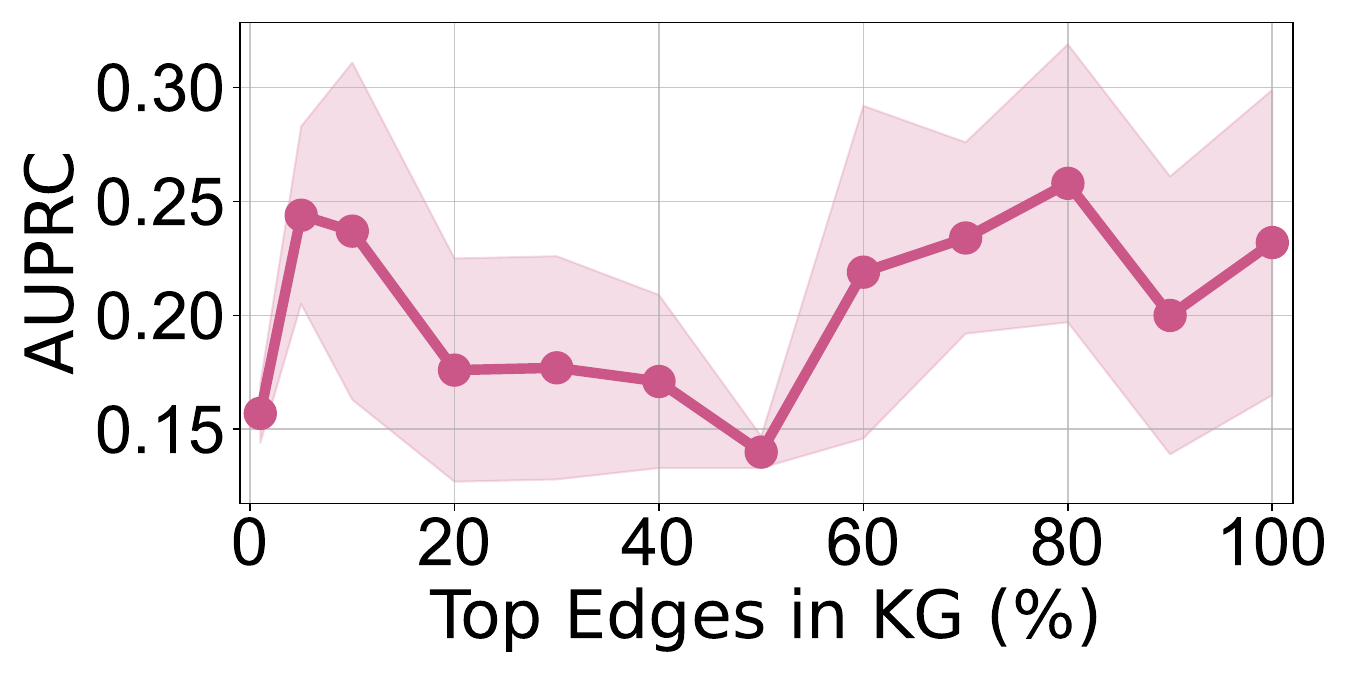}
        \vspace{0.1cm}
        \end{subfigure}
    \end{minipage}
    \hfill
    \begin{minipage}[t]{0.32\textwidth}
        \centering
        \setcounter{subfigure}{10}
        \begin{subfigure}[t]{\textwidth}
        \caption*{90d Mortality (MIMIC)}
        \vspace{-0.1cm}
        \includegraphics[width=0.91\textwidth]{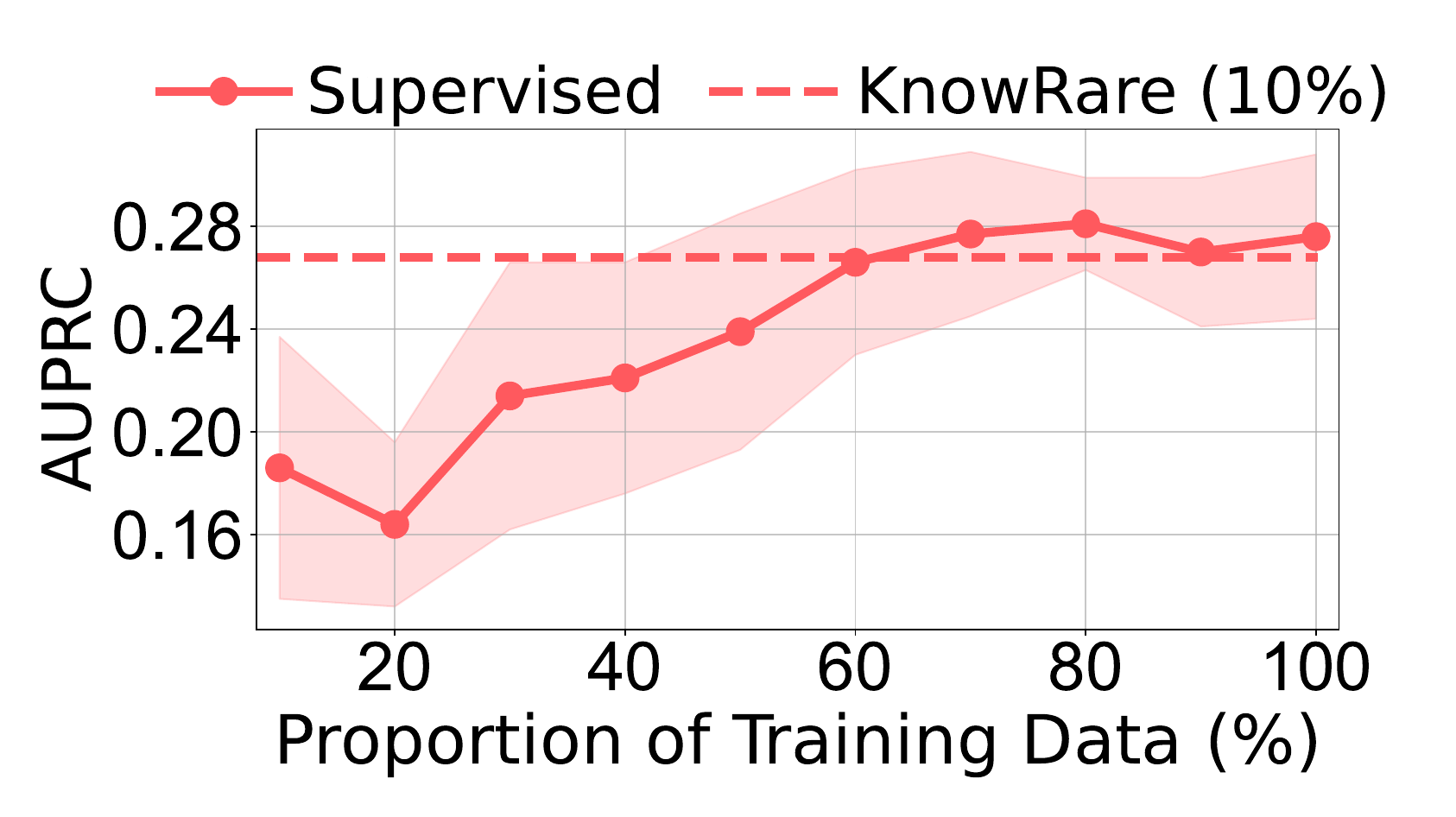}
        \vspace{0.1cm}
        \end{subfigure}
        
        \begin{subfigure}[t]{\textwidth}
        \caption*{30d Readmission (MIMIC)}
         \vspace{-0.1cm}
            \includegraphics[width=0.91\textwidth]{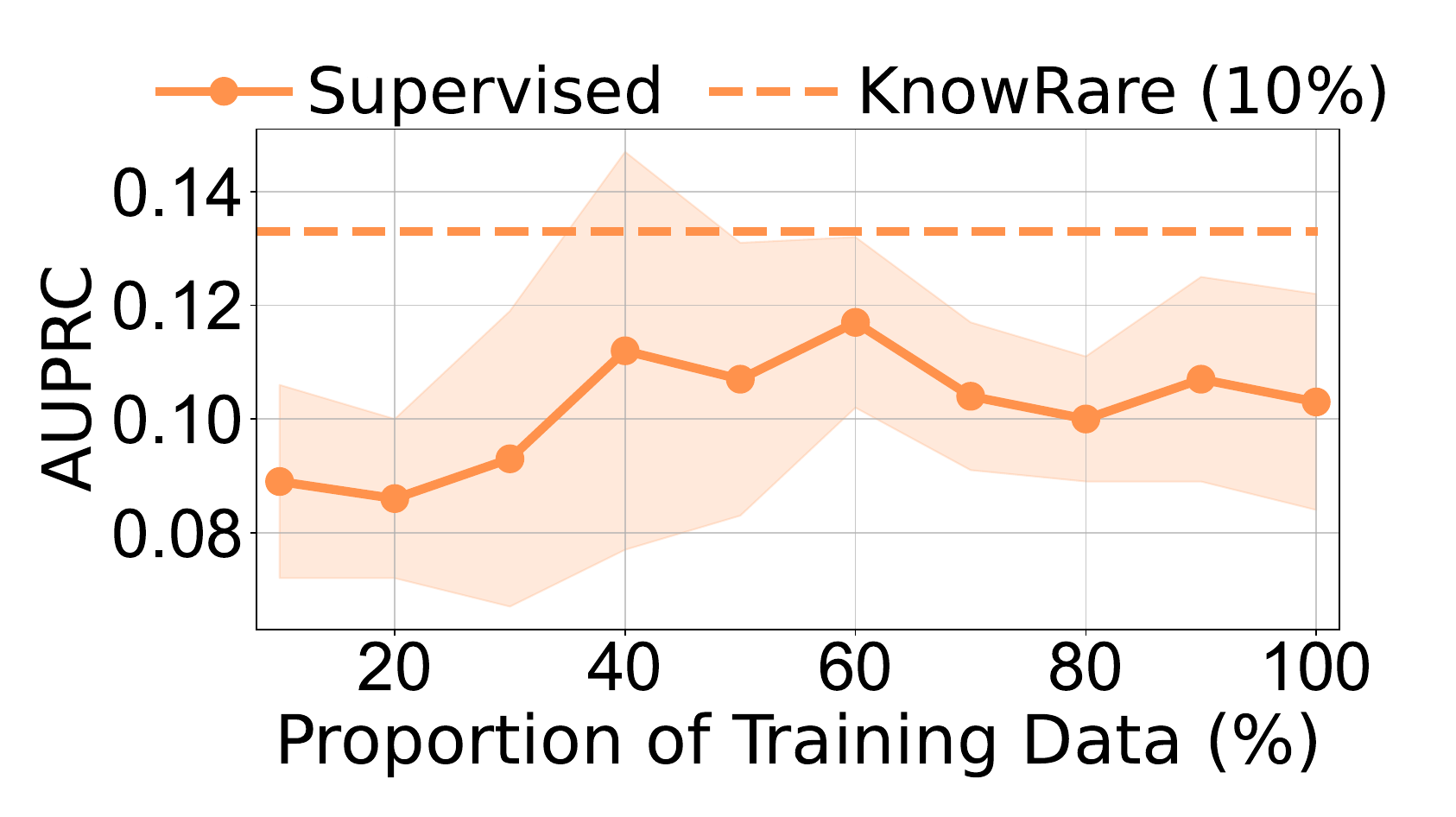} 
        \vspace{0.1cm}
        \end{subfigure}
        
        \begin{subfigure}[t]{\textwidth}
        \caption*{ICU Mortality (eICU)}
        \vspace{-0.1cm}
        \includegraphics[width=0.91\textwidth]{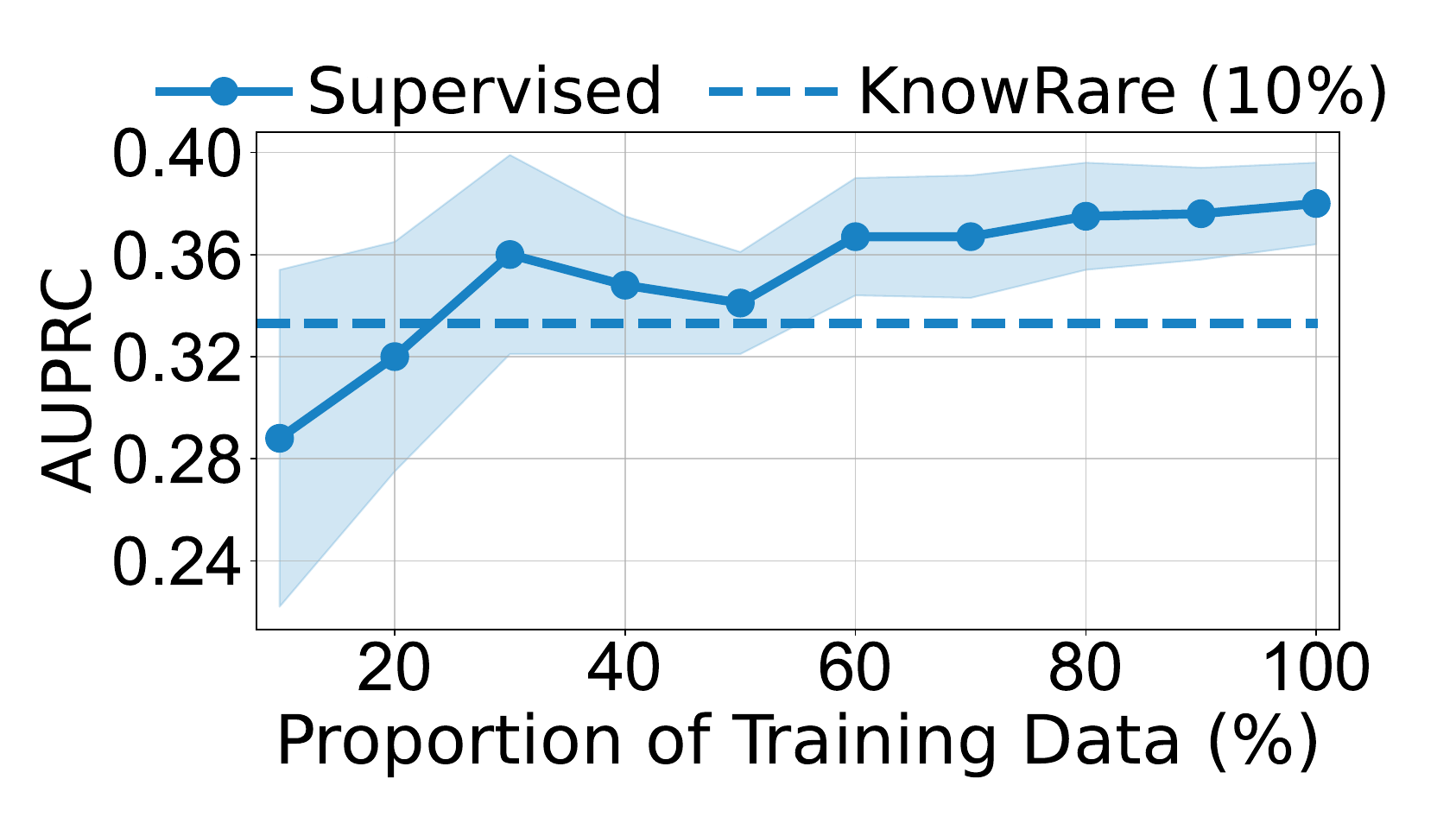}
        \vspace{0.1cm}
        \end{subfigure}
        
        \begin{subfigure}[t]{\textwidth}
            \caption*{LoS (eICU)}
            \vspace{-0.1cm}
            \includegraphics[width=0.92\textwidth]{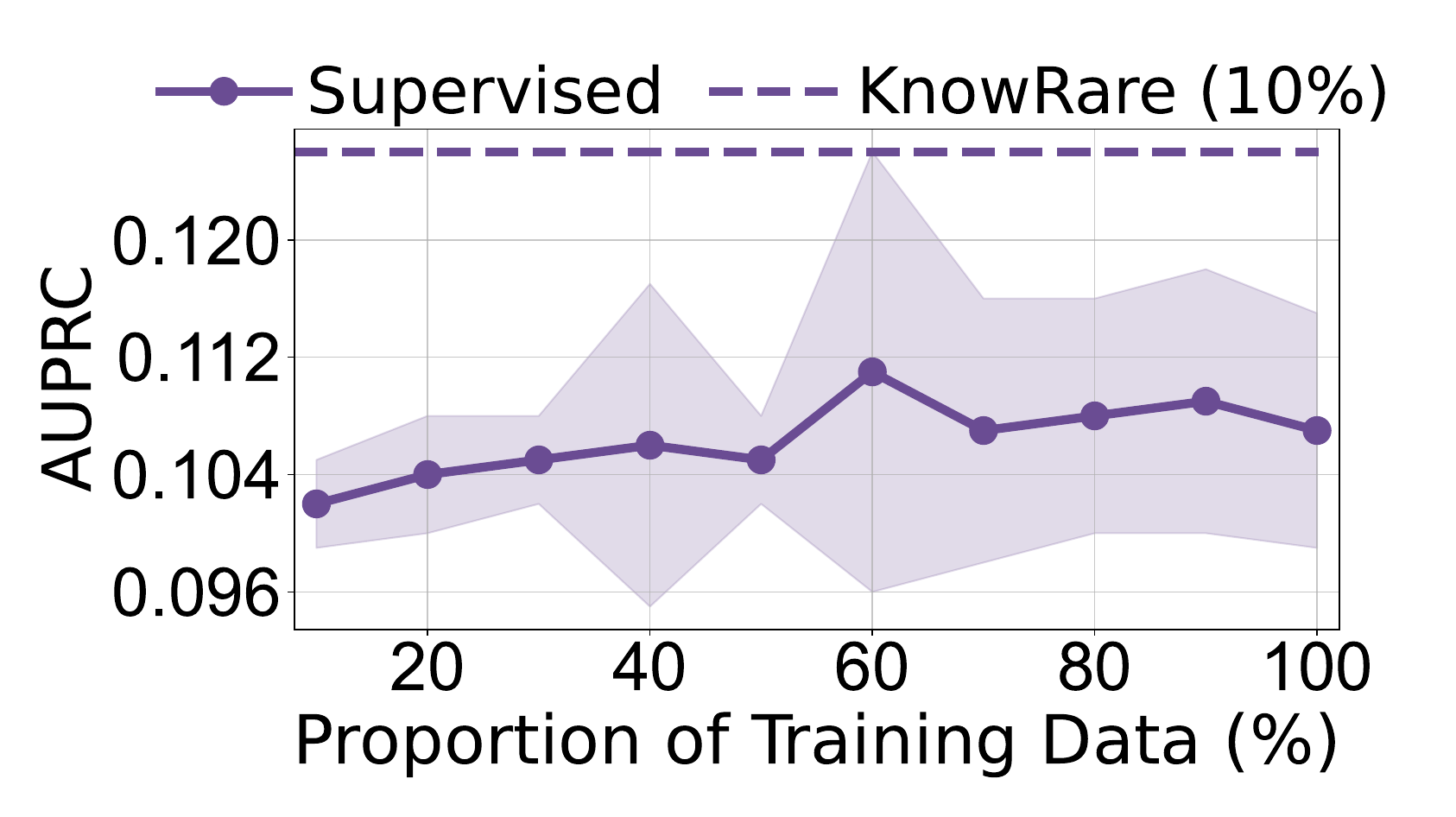}
        \vspace{0.1cm}
        \end{subfigure}
        
        \begin{subfigure}[t]{\textwidth}
            \caption*{Phenotyping (eICU)}
            \vspace{-0.1cm}
            \includegraphics[width=0.92\textwidth]{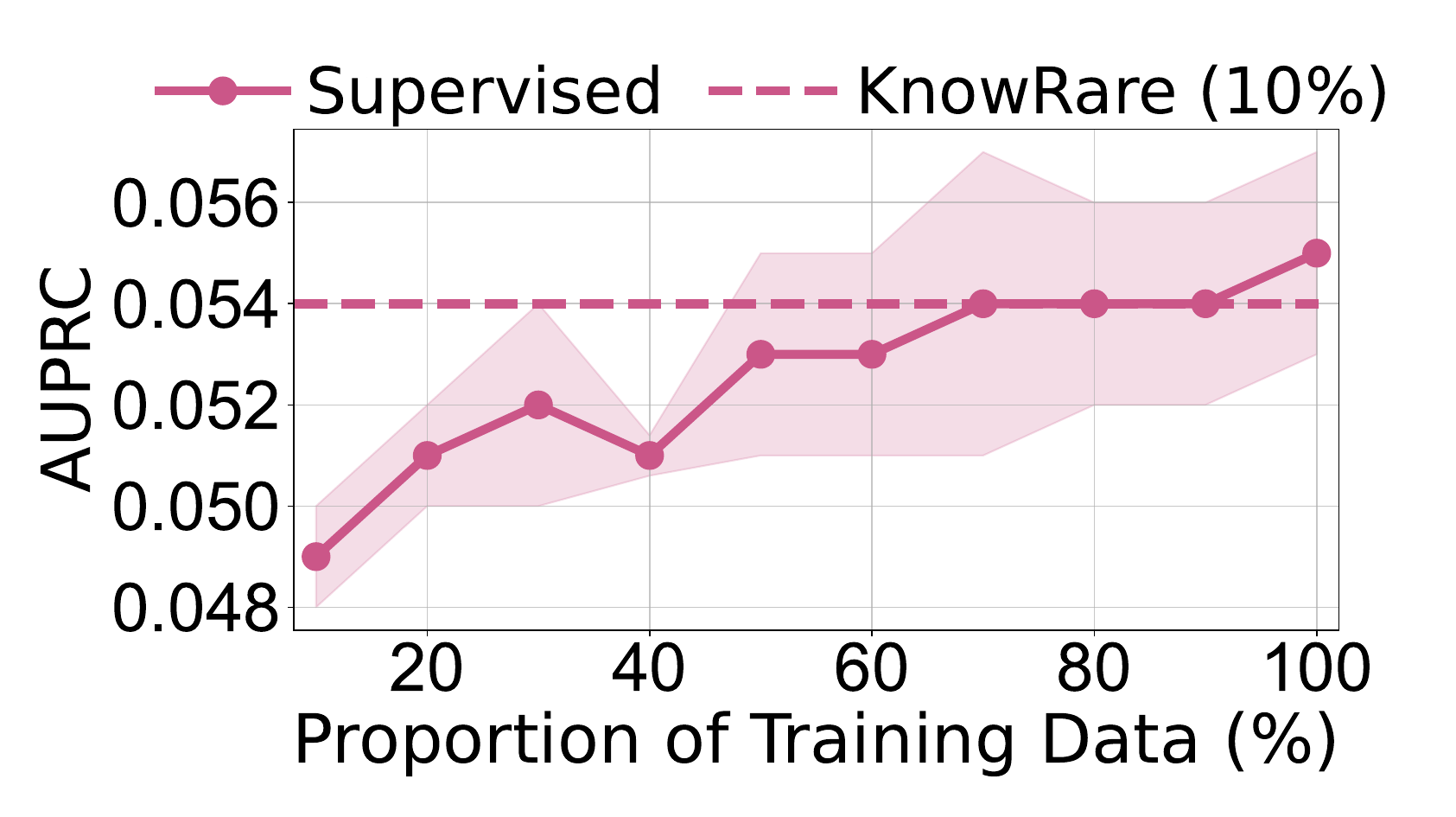}
        \vspace{0.1cm}
        \end{subfigure}
    \end{minipage}
    
    \caption{\textbf{Case studies: Evaluating KnowRare's adaptability and generalisation in clinical practice.} Studies include: \textbf{(a)} Analysis of required source conditions: Assessing KnowRare's adaptability to different hospital datasets and clinical prediction tasks by varying the proportion of source conditions included from 1\% to 100\%. \textbf{(b)} Impact of condition KG sparsity: Assessing KnowRare's sensitivity and adaptability to varying KG completeness by retaining different proportions of top-weighted edges (from 1\% to 100\%) in the KG. \textbf{(c)} Generalisation to common conditions under limited-data scenarios: Evaluating KnowRare's robustness by training it with only 10\% of the available septicemia data, while the LSTM baseline model uses septicemia data ranging from 10\% to 100\%. Rows correspond to the five prediction tasks: (1) 90-day mortality prediction after hospital discharge (MIMIC-III), (2) 30-day readmission prediction after hospital discharge (MIMIC-III), (3) ICU mortality prediction (eICU), (4) Remaining length of stay prediction (eICU), and (5) Phenotyping prediction (eICU). Points represent mean values, and shaded regions indicate standard deviation over five runs.}
    \label{fig:case studies}
\end{figure}

\subsection*{Case Studies: Evaluation of KnowRare's Adaptability and Generalisation}

This section presents three case studies designed to evaluate KnowRare's adaptability and generalisation in real-world clinical settings. Specifically, we conducted experiments to assess: (1) the number of source conditions required by KnowRare to adapt effectively across different hospital datasets and prediction tasks; (2) the impact of condition KG sparsity on KnowRare's adaptability to different dataset characteristics; and (3) KnowRare's ability to generalise robustly to common conditions under limited training data scenarios.

To evaluate the influence of source condition diversity on domain adaptation, we varied the proportion of source conditions selected for KnowRare training, ranging from 1\% to 100\% of the available source conditions. The objective was to determine whether the broader inclusion of source conditions enhances generalisation or introduces confounding. The results, illustrated in Figure~\ref{fig:case studies}(a), reveal a non-linear relationship between the quantity of the source condition and the performance of the model. Initially, increasing the proportion of source conditions improves the accuracy of the task, with peak performance occurring at 10–20\% across tasks. Beyond this threshold, performance declines sharply for ICU mortality, LoS, and phenotyping predictions in the multi-centre setting, eventually stabilising at higher proportions. In particular, for LoS prediction, minimal source diversity (1\% of conditions) yields superior results compared to larger selections.

In addition, we investigated the effect of the completeness of the condition KG. This was achieved by iteratively retaining only the top n\% of the weighted edges (ranging from 1\% to 100\%) in the KG and re-evaluating the KnowRare framework. The experiment aimed to identify whether sparser, high-confidence relationships or denser, inclusive graphs better support adaptation in clinical prediction tasks. As shown in Figure~\ref{fig:case studies}(b), the optimal proportion of retained edges differs markedly between datasets. For the multi-centre eICU cohort, performance peaks when the top 5\% of KG edges are included, except for the phenotyping prediction task. The single-centre MIMIC-III dataset achieves maximal AUPRC with 60–70\% of edges. Beyond these thresholds, inclusion of lower-weighted edges correlates with progressive performance degradation. One outlier appears at the phenotyping prediction task, which exhibits a dual-peak pattern in the eICU dataset, achieving optimal performance at both 5\% and 80\% edge retention.

We further evaluated KnowRare's applicability to common conditions in scenarios characterised by limited training data. Common conditions can also experience data scarcity due to practical constraints. To examine whether KnowRare could generalise to such situations, we experimented with septicaemia, a common ICU condition, deliberately restricting the training set to only 10\% of the available samples. Importantly, the condition-agnostic pre-training module was disabled in this experiment to rigorously prevent data leakage, thus ensuring an accurate evaluation of KnowRare's ability to exploit general clinical knowledge when faced with severe resource constraints. Our results (Figure~\ref{fig:case studies}(c)) demonstrated that KnowRare achieved comparable or superior performance to a standard LSTM model trained on all available data on 90-day mortality, 30-day readmission, and remaining LoS. In ICU mortality and phenotyping prediction tasks, the standard LSTM model only surpassed KnowRare's performance when trained with at least three times more labelled data. These findings underscored KnowRare's potential to effectively leverage clinical insights from similar conditions, thereby improving predictive performance in data-limited clinical scenarios.

\subsection*{Explanability Analysis}

\begin{figure}[ht]
    \centering
    \includegraphics[width=0.8\linewidth]{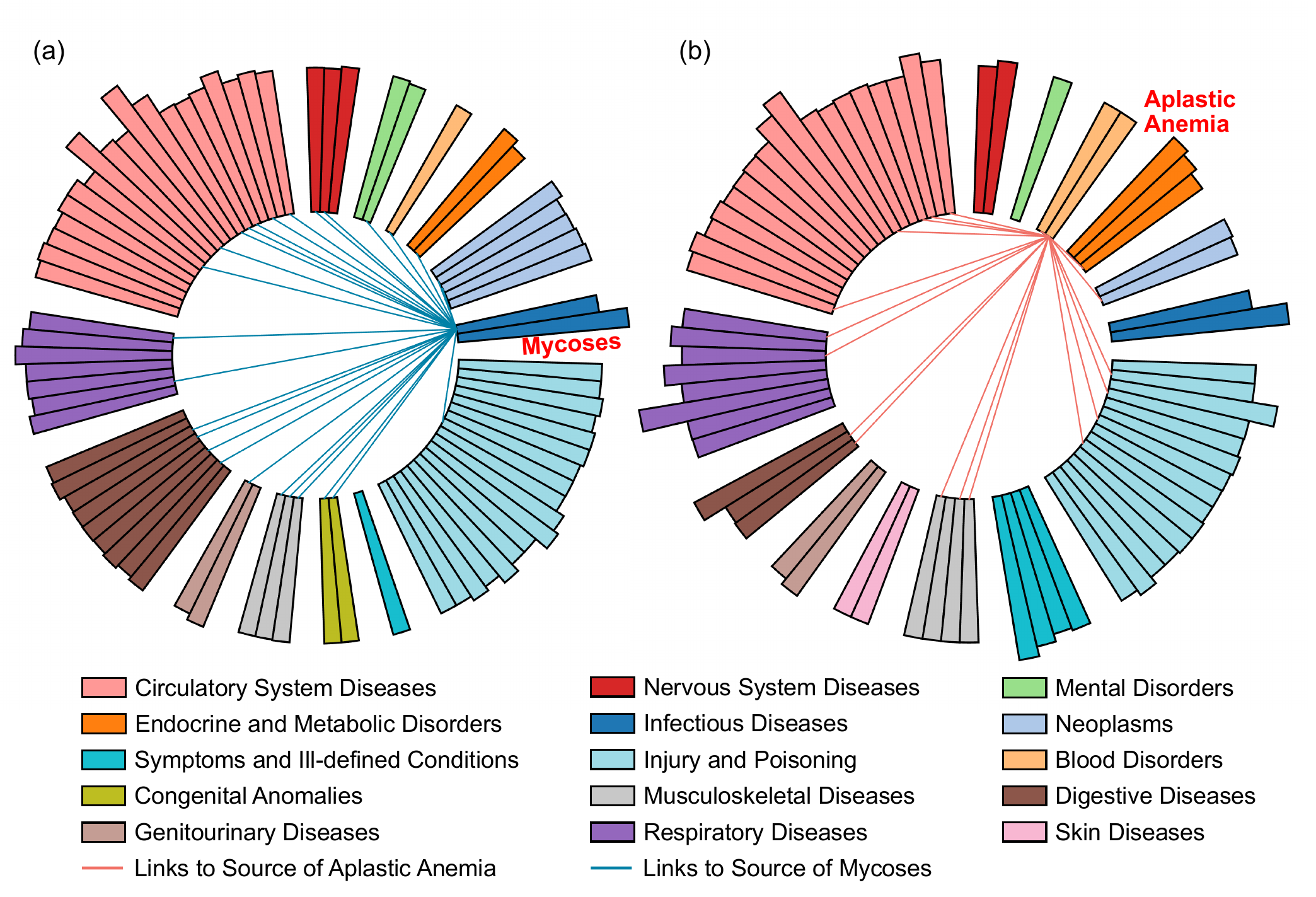}
    \vspace{-6px}
    \caption{\textbf{Visualisation of conditions selected by KnowRare as source conditions for predicting mycoses in MIMIC-III and aplastic anaemia (AA) in eICU.} Conditions are categorised according to the ICD-9-CM classification system, with each category represented by a distinct colour. The height of each bar corresponds to the number of patients diagnosed with each condition, indicating their relative prevalence within the datasets. This visualisation illustrates how KnowRare selects clinically similar source conditions different from ICD-9-CM hierarchical relationships.}
    \label{fig:icd_visualisations}
\end{figure}

To evaluate the explanability of KnowRare's selection of source conditions, we visualised its choices for two rare conditions: mycoses, a low-prevalence condition in MIMIC-III, and aplastic anaemia (AA), a rare recognised condition in the eICU. For each case, we included the 40 most prevalent conditions, the 10 least prevalent conditions, and the subset of conditions selected by KnowRare (Figure~\ref{fig:icd_visualisations}). This analysis aimed to determine whether the framework prioritises hierarchy from the ICD-9-CM coding system or instead identifies clinical similarities directly from the heterogeneous EHR data.

The analysis yielded two key observations. First, more than 95\% of the source conditions selected by KnowRare belonged to ICD-9-CM categories distinct from those of the target rare condition. Specifically, for mycoses, KnowRare predominantly selected source conditions from circulatory system disorders (ICD-9-CM 390–459), even though fungal infections are classified under infectious conditions (ICD-9-CM 001–139). Likewise, for AA, the model primarily selected source conditions from injury and poisoning (ICD-9-CM 800–999), despite AA being categorised as a blood disorder (ICD-9-CM 280–289). These results indicated that KnowRare captured clinical similarities directly from heterogeneous patient data rather than relying solely on established ICD-9-CM hierarchies. 

\section*{Discussion}

In this study, we proposed KnowRare, a DL framework designed to improve the prediction of clinical outcomes for rare conditions in the ICU. KnowRare addresses two critical challenges, data scarcity and intra-condition heterogeneity, by combining condition-agnostic pre-training with knowledge-guided domain adaptation. Evaluation of two publicly available ICU datasets (MIMIC-III and eICU) demonstrated that KnowRare consistently outperformed baseline methods, achieving improvements of up to 17.0\% in AUPRC. The consistent improvement across multiple clinical tasks, including 90-day mortality after discharge, 30-day readmission after discharge, ICU mortality, ICU LoS, and phenotyping, demonstrates KnowRare's potential to improve clinical decision-making for rare conditions in ICUs.

We observed a clear "data-volume paradox" with standard methods in our experiments (Table~\ref{table:comparison}): training exclusively on target rare conditions resulted in significantly inferior performance, whereas aggregating data from multiple conditions improved results in single-centre scenarios but degraded performance in multi-centre scenarios. This paradox underscores the necessity for careful selection of aggregated data sources to ensure that increased data volume does not compromise the quality and clinical relevance of training data. KnowRare addresses this issue through knowledge-guided domain adaptation, selectively identifying and utilising only the most similar source conditions. This targeted approach enhances data volume without amplifying irrelevant noise, contributing significantly to KnowRare's effectiveness.

Specialised methods for rare conditions demonstrated strong performance in individual prediction tasks, but lacked robustness when evaluated across multiple clinical outcomes (Table~\ref{table:comparison}). In contrast, KnowRare showed consistent effectiveness across various clinical tasks. This improved generalisation was primarily driven by its two-stage training strategy: initially capturing generalisable temporal patterns through condition-agnostic pre-training, followed by fine-tuning with knowledge-guided domain adaptation to selectively activate condition-specific knowledge for each rare condition. This finding is consistent with related research for cases with limited data~\cite{zhang2019metapred, zhao2024leave}. Specifically, the condition-agnostic pre-training process enables KnowRare to establish robust baseline representations capable of generalising effectively to data-scarce conditions. Furthermore, the knowledge-guided domain adaptation, particularly through the domain selection module, substantially addresses intra-condition heterogeneity (Table~\ref{tab:ablation}). By selecting clinically similar conditions based on the condition KG, this module exposes KnowRare to diverse clinical patterns without introducing excessive noise. Consequently, KnowRare effectively addressed the challenges of data scarcity and intra-condition heterogeneity of rare conditions in the ICature

For the most critical ICU mortality prediction, KnowRare outperformed traditional ICU scoring systems, including APACHE IV and IV-a. These systems typically rely on large cohorts that primarily comprise common conditions, which limits their effectiveness for rare conditions. In contrast, KnowRare's tailored selection of clinically similar conditions and knowledge-guided adaptation processes enabled superior performance specifically for rare conditions.

The selection of source conditions by KnowRare was primarily based on data-driven relationships rather than strictly following the ICD coding hierarchies (Figure~\ref{fig:icd_visualisations}). This approach is particularly effective for rare conditions, where standard ICD categories often do not capture nuanced cohort-specific similarities. By identifying clinically meaningful relationships, KnowRare not only improves the interpretability of the model, but also provides clinicians with valuable insights into the similarities of the conditions. This information can help clinicians better understand how KnowRare generates predictions and leverages knowledge from clinically similar conditions, enhancing trust and acceptance of its predictions.

Interestingly, our finding contradicted the conventional idea that using more data leads to better performance until saturation \cite{mahmood2022much, zhang2021you}. Instead, KnowRare demonstrated optimal domain adaptation when using approximately 10\% of the most similar source conditions. Including more data introduced excessive noise, hindering effective domain adaptation. Particularly in the multi-centre eICU dataset, significant performance degradation occurred beyond 50\% data inclusion (Figure~\ref{fig:case studies}(a)). However, an exception was the length of stay (LoS) prediction task, where restricting data to only 1\% of the most similar conditions optimally prevented overfitting to irrelevant clinical patterns. 

Our findings also show KnowRare's capability to adapt to diverse ICU scenarios by adjusting the retention threshold of relationships in the condition KG (Figure~\ref{fig:case studies}(b)). Optimal performance in multi-centre datasets (eICU) was achieved by retaining the strongest 5\% of the graph edges, while single-centre datasets (MIMIC-III) benefited from the retention of a higher proportion (60–70\%). An exception emerged in phenotyping prediction, which exhibited a dual-peak performance pattern. Phenotyping prediction benefited from both highly confident, tightly coupled clinical relationships at lower retention thresholds and broader, weaker connections at higher thresholds. This unique pattern can be caused by the distinct nature of phenotyping prediction, where multiple concurrent patterns must be identified simultaneously~\cite{harutyunyan2019multitask}. Besides the dataset-specific threshold, subtle, long-range clinical relationships can enhance the identification of meaningful phenotypes overlooked by strong connections, therefore contributing to better performance in phenotyping prediction. 

Beyond rare conditions, KnowRare extended its utility to common conditions that experienced data scarcity. Our evaluation with limited training samples for septicaemia demonstrated that KnowRare could achieve comparable or superior predictive performance compared to standard DL models trained on all data for the common condition. This result highlights KnowRare's potential utility in clinical settings facing real-world data constraints due to operational or ethical limitations \cite{li2024exploring, van2014systematic}.

Several limitations of our study should be acknowledged. Firstly, our reliance on ICD-9 coding constrained the granularity of condition categorisation, although this limitation arose directly from the datasets used (MIMIC-III and eICU). This limited granularity could reduce the accuracy of capturing subtle clinical distinctions among rare conditions, potentially limiting the effectiveness of knowledge-guided domain adaptation. Secondly, differences in ICD-CM availability restricted our ability to evaluate identical tasks across datasets. Specifically, the ICD-CM codes in MIMIC-III, available only at discharge, limited task choices to post-discharge outcomes. This limitation potentially affected the generalisability of observations across datasets.

In conclusion, KnowRare effectively bridges data gaps to predict the outcomes of rare conditions in the ICU by integrating general knowledge and selectively adapting insights from clinically similar conditions. Validated through extensive experiments on real-world ICU datasets, KnowRare demonstrates robust predictive performance and strong potential to support clinical decision-making. Future research should focus on prospective validation in clinical settings and investigate opportunities for integration within broader healthcare systems to further enhance care for rare ICU conditions.

\section*{Methods}
\subsection*{Data processing}
\label{sec:data processing}

To ensure a comprehensive representation of patient data across the datasets, we extracted variables, including demographic data, vital signs, and laboratory tests from MIMIC-III and eICU, as shown in the Supplementary Table~\ref{table:variables}. The extraction process involves multiple steps, including cohort selection, variable aggregation, missing value imputation, and normalisation. First, we exclude patients with a hospital stay of less than 48 hours in the MIMIC-III dataset and those with an ICU stay of less than 24 hours in the eICU dataset to ensure sufficient data availability for inference. Then, each patient is assigned a primary diagnosis based on the ICD-9-CM coding system, using the first three levels of their ICD-9-CM code. To maintain a sufficient sample size for a robust evaluation, we exclude conditions with fewer than ten patients. After applying these criteria, the final dataset consists of 38,360 MIMIC-III samples and 72,536 eICU samples. We then split each dataset into training sets (67\%), validation sets (16\%), and test sets (17\%) at the patient level. Specifically, we stratify the patients for each condition across splits to ensure a representative distribution in the training, validation, and test sets. After selecting the relevant patient cohorts, we extract vital signs and laboratory test measurements from structured EHR tables. For MIMIC-III, we select the first recorded values from the last 48 hours of hospital admission, while for eICU, we extract the recorded values from the first 24 hours of ICU admission. Demographic variables such as age, gender, and race are obtained directly from patient metadata. To create a structured time-series representation, we aggregated the extracted variables into fixed time resolutions following established benchmarks \cite{gupta2022extensive, harutyunyan2019multitask}. For MIMIC-III, we segment the time-series data into 2-hour windows and compute the mean value within each window, resulting in 24 time steps spanning the last 48 hours of hospital admission. For eICU, we apply a similar method using 1-hour windows, also producing 24 time steps covering the first 24 hours of ICU admission. If no recorded values are available within a given time step, they are left as missing values at this stage. To handle missing values, we follow widely adopted imputation strategies \cite{van2023yet, hyland2020early}. First, we apply forward imputation, where missing values are replaced with the last available measurement for the same patient. If no previous value exists, we apply backwards imputation, filling in missing values using the next available measurement from the same patient. After these steps, any remaining missing values are imputed using the mean value of the corresponding variables computed from the training set. This ensures that the validation and test sets remain independent and do not incorporate statistical information from unseen data. Finally, to standardise the variable scales, we normalise all continuous variables using Z-score normalisation, where each variable is transformed using the mean and standard deviation computed from the training set. This normalisation step ensures consistent variable distributions across different datasets and tasks while preventing data leakage.

\subsection*{Baseline Models}

To comprehensively evaluate KnowRare, we compared its performance against two categories of baseline methods: standard models commonly used for clinical prediction tasks and specialised methods explicitly designed for data-scarce scenarios. Detailed descriptions of the baseline methods are summarised in Table~\ref{tab:baseline_methods}.

\begin{table}[ht!]
\centering
\caption{Baseline methods included in this study.}
\vspace{-8px}
\label{tab:baseline_methods}
\begin{tabular}{ll}
\toprule
\textbf{Name}        & \textbf{Description}   \\
\midrule
LSTM \cite{hochreiter1997long} & Long short-term memory network for modelling EHR temporal dependencies.  \\
Transformer \cite{vaswani2017attention} & Self-attention architecture capturing long-range variable interactions.  \\
RETAIN \cite{choi2016retain} & Interpretable attention-based recurrent neural network for clinical time-series. \\
MetaPred \cite{zhang2019metapred} & Gradient-based meta-learning framework for few-shot condition diagnosis. \\
RareMed \cite{zhao2024leave} & Pre-training on clinical notes to enhance rare condition medication recommendations. \\
SMART \cite{yu2024smart} & Self-supervised learning via masked reconstruction of EHR data. \\
FADA \cite{teshima2020few} & Few-shot classification via adversarial domain alignment. \\
AdvDiag \cite{zhang2022adadiag} & Adversarial training to handle cross-population diagnostic distribution shifts. \\
Stable-CRP \cite{lee2023stable} & Patient data reweighting for stable predictions amidst temporal shifts. \\
MANYDG \cite{yang2023manydg} & Learning domain-invariant representations to generalise across patient domains.  \\
\bottomrule
\end{tabular}
\end{table}

\subsection*{The KnowRare Framework}\label{KnowRare}
\subsubsection*{Condition Knowledge Graph Construction.} 
\label{sec:Knowledge Graph Construction}

To identify clinically similar source conditions, a heterogeneous {condition KG} $\mathcal{G} = (\mathcal{V}, \mathcal{E})$ is constructed from the EHR database, where $\mathcal{V}$ represents the set of conditions, $\mathcal{E}$ defines the types of relationships that capture similarities of the condition. The graph encodes three types of relationships to model condition similarities.

\paragraph{\textbf{Diagnosis Similarity.}}  
The diagnosis-based relation captures co-occurrence patterns between conditions within diagnosis records. In EHRs, multiple ICD-CM codes are assigned during a patient visit, resulting in frequent co-occurrence of conditions within the same diagnosis record. A triplet $(v_i, r_1, v_j) \in \mathcal{E}$ is established between conditions $v_i, v_j \in \mathcal{V}$ if they frequently appear together in patient records. The edge weight is computed as the normalised co-occurrence frequency:
\begin{equation} 
w_{ij}^{(r_1)} = \frac{\text{CoOcc}(v_i, v_j)}{\sum_k \text{CoOcc}(v_i, v_k)},
\end{equation}
where $\text{CoOcc}(v_i, v_j)$ represents the number of times conditions $v_i$ and $v_j$ appear together in the same diagnosis record.

\paragraph{\textbf{Record Similarity.}}  
The record-based relation models condition similarity based on statistical patterns in patient variables. Each condition $v \in \mathcal{V}$ is represented by a characteristic vector $\mathbf{s}_v$, calculated as the mean and standard deviation of the patient-level variables across all patients diagnosed with the condition:
\begin{equation} 
\mathbf{s}_v = \left[\text{mean}(\mathcal{X}_v), \text{std}(\mathcal{X}_v)\right],
\end{equation}
where $\mathcal{X}_v = \{\mathbf{X}_p \mid p \in \mathcal{P}_v\}$ denotes the set of time-series variables of all patients $\mathcal{P}_v$ diagnosed with condition $v$. The mean and standard deviation are computed element-wise across all patient records.

The similarity weight between conditions $v_i$ and $v_j$ is computed using the inverse L2 distance:
\begin{equation} 
w_{ij}^{(r_2)} = \frac{1}{1 + \|\mathbf{s}_{v_i} - \mathbf{s}_{v_j}\|_2}.
\end{equation}

To retain meaningful relationships, only the top 50\% of the highest-weighted record-based connections are preserved.

\paragraph{\textbf{Drug Similarity.}}  
The drug-based relationship captures the similarity of the conditions based on the use of shared medications. For each condition $v \in \mathcal{V}$, let $\mathcal{A}_v$ denote the set of drugs administered to the patients diagnosed with that condition. The weight of the relationship between conditions $v_i$ and $v_j$ is calculated using the Jaccard similarity of their drug sets:
\begin{equation}
w_{ij}^{(r_3)} = \frac{|\mathcal{D}_{v_i} \cap \mathcal{D}_{v_j}|}{|\mathcal{D}_{v_i} \cup \mathcal{D}_{v_j}|}.
\end{equation}

Similarly, only the top 50\% of the highest-weighted drug-based relations are retained.

\subsubsection*{Knowledge-guided domain selection}

After constructing the condition KG $\mathcal{G} = (\mathcal{V}, \mathcal{E})$ (Section~\ref{sec:Knowledge Graph Construction}), condition relationships are embedded into a shared latent space using a KG embedding model. In this study, we adopt the TuckER model \cite{balazevic2019tucker}. TuckER factorises the KG tensor $\mathcal{T} \in \mathbb{R}^{|\mathcal{V}| \times |\mathcal{R}| \times |\mathcal{V}|}$ into a shared core tensor $\mathbf{W} \in \mathbb{R}^{d \times d \times d}$ and the embeddings of the relation / entity. The score for a triple $(v_i, r_k, v_j)$ is computed as
\begin{equation}
\mathbf{T}_{ilj} = \mathbf{W} \times_1 \mathbf{E}_{v_i} \times_2 \mathbf{R}_{r_k} \times_3 \mathbf{E}_{v_j},
\label{eq:tucker}
\end{equation}
where $\mathbf{E}_{v_i}, \mathbf{E}_{v_j} \in \mathbb{R}^{d}$ represent the embeddings of conditions $v_i$ and $v_j$, respectively, $\mathbf{R}_{r_k} \in \mathbb{R}^{d}$ denotes the embedding of relation $r_k$, and $\times_n$ indicates the mode-$n$ tensor product.

For a target rare condition $v_t$, the most similar source conditions are selected based on the cosine similarity of the condition embeddings:
\begin{equation} 
\mathcal{S}^{*} = \arg\max_{\substack{\mathcal{S}^{*} \subseteq \mathcal{S} \\ |\mathcal{S}^{*}| = k}} \sum_{v_s \in \mathcal{S}^{*}} \frac{\mathbf{E}_{v_t} \cdot \mathbf{E}_{v_s}}{\|\mathbf{E}_{v_t}\| \|\mathbf{E}_{v_s}\|}.
\label{eq:condition selection}
\end{equation}

The top-$k$ conditions with the highest similarity scores are selected as source domains for adaptation.

\subsubsection*{Condition-Agnostic Pre-training} \label{sec:Pretraining}
We develop KnowRare using an LSTM network \cite{hochreiter1997long} as the backbone for time-series encoding, given its proven effectiveness in capturing temporal dependencies in sequential EHR data \cite{men2021multi}. To capture general temporal and contextual representations from EHR data, the proposed KnowRare framework includes a condition-agnostic pre-training stage, ensuring that the model learns generalisable patterns without overfitting to any specific task. Specifically, we employ a self-supervised method based on next-step prediction.

The encoder consists of two separate modules: a temporal encoder $f^{\text{temp}}$ for time-series variables $\mathbf{X}$ and a contextual encoder $f^{\text{cont}}$ for contextual variables $\mathbf{C}$. The encoded variables are concatenated and mapped to obtain the latent representation:
\begin{equation}
\mathbf{h}_t = f^{\text{proj}} \left( f^{\text{temp}} (\mathbf{x}_t), f^{\text{cont}} (\mathbf{C}) \right),
\end{equation}
where $\mathbf{h}_t \in \mathbb{R}^{d_h}$ is the hidden representation at time $t$.

By self-supervised learning, the encoder extracts general latent representations that can be quickly adapted to rare conditions. The pre-training objective is designed as a multivariate trajectory reconstruction task, where a decoder \( f^{\text{dec}} \) predicts the next timestep conditioned on the fused representation:  

\begin{equation}
\mathcal{L}_{\text{pre-train}} = \frac{1}{T-1} \sum_{t=1}^{T-1} \|\mathbf{X}_{t+1} - f^{\text{dec}} (\mathbf{h}_t)\|^2.
\label{eq:pretrain}
\end{equation} 

By disentangling latent variables with task-specific patterns, this stage learns condition-agnostic representations that encode general knowledge as the initial parameters for target condition adaptation. This stage is valuable in dealing with data scarcity.

\begin{algorithm}[ht!]
\caption{Training Process of KnowRare}
\label{alg:knowrare}
\begin{algorithmic}[1]

\Require 
EHR data $\{(\mathbf{X}_p, \mathbf{C}_p, y_p)\}$, 
condition KG $\mathcal{G} = (\mathcal{V}, \mathcal{E})$, 
rare condition $v_t$, 
hyperparameters $k, \lambda$
\Ensure
Trained model parameters $\theta^{*}$

\State \textbf{Step 1: Condition-agnostic Pre-training}
\State Initialise encoder parameters $\theta_0$
\For{each patient $p$}
    \For{$t = 1$ to $T-1$}
        \State $\mathbf{h}_t \gets f^{\text{proj}}\bigl(f^{\text{temp}}(\mathbf{x}_t), f^{\text{cont}}(\mathbf{C}_p)\bigr)$
        \State $\hat{\mathbf{X}}_{t+1} \gets f^{\text{dec}}(\mathbf{h}_t)$
        \State Update $\theta_0$ with Equation~\eqref{eq:pretrain}
    \EndFor
\EndFor
\State \textbf{Step 2: Knowledge-guided Domain Adaptation}
\State Obtain embeddings $\{\mathbf{E}_{v}\}$ from $\mathcal{G}$ with Equation~\eqref{eq:tucker}

\State Select $\mathcal{S}^*$ of $k$ conditions based on $\{\mathbf{E}_{v}\}$ with Equation~\eqref{eq:condition selection}
\State Initialise $\theta \gets \theta_0$, define discriminator $d_\phi$
\While{not converged}
    \State Sample mini-batches from $\mathcal{S}^*$ and $\mathcal{D}_t$
    \State $\mathbf{h}_T \gets f^{\text{proj}}\bigl(f^{\text{temp}}(\mathbf{X}),f^{\text{cont}}(\mathbf{C})\bigr)$
    \State $\hat{y} \gets f_\theta(\mathbf{h}_T)$
    \State Compute losses $\mathcal{L}_{\text{pred}}, \mathcal{L}_{\text{adv}}$
    \State Update $\theta$ and $\phi$ with Equation~\eqref{eq:adversarial training}
\EndWhile

\State \textbf{return} Model parameters $\theta^*$ for $v_t$

\end{algorithmic}
\end{algorithm}

\subsubsection*{Joint Adversarial Domain Adaptation} \label{sec:Joint Adversarial Learning}

The final stage of the KnowRare framework employs adversarial learning to align condition-level distributions for robust adaptation to the target rare condition. Leveraging the final time-step latent representation $\mathbf{h}_T$ that encodes the full temporal dynamics of the input variables, we mitigate the difference in the joint distribution of the latent representation and the prediction. Unlike prior works that align marginal distributions \cite{zhang2022adadiag, lee2023stable}, we hypothesise that domain shifts arise from discrepancies in the joint distribution $P(\mathbf{h}_T, y)$ of both representations and clinical outcomes. To address this, we propose a joint adversarial domain adaptation that considers both latent representations and task-specific predictions. Specifically, a joint discriminator $d_\phi$ operates on the concatenation of $\mathbf{h}_T$ and predicted outcome $\hat{y}$, aligning the variance in the joint distribution:
\begin{equation}
\mathcal{L}_{\text{adv}} = -\mathbb{E}_{(\mathbf{h}_T, \hat{y})} \sum_{i=1}^{|\mathcal{S}^*|+1} y_{\mathcal{D}_i} \log d_\phi(\mathbf{h}_T, \hat{y}),
\end{equation}
where $y_{\mathcal{D}_i} \in \mathcal{V}$ denotes the domain labels. 

With the discriminator and the adversarial loss, we induce a minimax optimisation process:  
\begin{equation}
\min_{\theta} \max_{\phi} \mathcal{L}_{\text{adv}}(\mathbf{h}_T, \hat{y}; \theta, \phi),
\label{eq:adversarial training}
\end{equation}
where $\theta$ and $\phi$ denote the encoder and discriminator parameters, respectively. Through this adversarial interaction, the encoder learns to extract domain-invariant representations, while the discriminator continuously refines its ability to distinguish domains. Eventually, the process converges to a Nash equilibrium, where the encoder produces latent representations that minimise domain discrepancy, facilitating effective domain adaptation.

The outcome prediction loss is computed via cross-entropy:
\begin{equation}
\mathcal{L}_{\text{pred}} =  \mathcal{L}_{\text{CE}}\left(f_\theta(\mathbf{X}, \mathbf{C}), y\right).
\end{equation}

To counteract data imbalance across conditions, we apply the inverse propensity score weighting, assigning each sample a weight $w_v = 1/p(v)$, where $p(v)$ is the prevalence of the condition $v$. This prioritises under-represented conditions during training.

The unified objective combines prediction and adversarial losses, with $\lambda$ balancing predictive performance and domain adaptation:
\begin{equation}
\mathcal{L}_{\text{total}} = \mathcal{L}_{\text{pred}} + \lambda \mathcal{L}_{\text{adv}}.
\end{equation}

For a clearer understanding, Algorithm~\ref{alg:knowrare} outlines the overall training process of KnowRare. The process contains two steps: (1) Condition-agnostic pre-training, where the encoder is pre-trained to learn generalisable temporal patterns; (2) Knowledge-guided domain adaptation, where the most similar source conditions are identified, and the model is trained with adversarial learning to align distributions across conditions.

\section*{Data Availability}
The MIMIC-III and eICU databases analysed in this study are available on the PhysioNet repositories \url{https://physionet.org/content/mimiciii/1.4/} and \url{https://physionet.org/content/eicu-crd/2.0/}.

\section*{Code Availability}
The code used for data processing and model development is available at \url{https://github.com/JasonZuu/KnowRare}.

\bibliography{sample}




\label{MainEnd}   

\clearpage        
\appendix

\makeatletter
  \let\OldThePage\thepage      
  \renewcommand{\thepage}{S\arabic{page}}
  \setcounter{page}{1}
\makeatother

\noindent
{\LARGE\bfseries Supplementary Information: Bridging Data Gaps of Rare Conditions in ICU: A Multi-Disease Adaptation Approach for Clinical Prediction}                

\setcounter{figure}{0}
\renewcommand{\thefigure}{S\arabic{figure}}

\setcounter{table}{0}
\renewcommand{\thetable}{S\arabic{table}}

\rfoot{\small\sffamily\bfseries\thepage/\pageref{SupEnd}}%

\section{Visualisation of Challenges of rare conditions in the ICU}

\begin{figure}[h!]
    \centering
    \captionsetup{justification=raggedright, singlelinecheck=false}
    \begin{subfigure}[b]{0.3\textwidth}
        \caption*{(a)}
        \vspace{-0.1cm}
        \includegraphics[width=\textwidth]{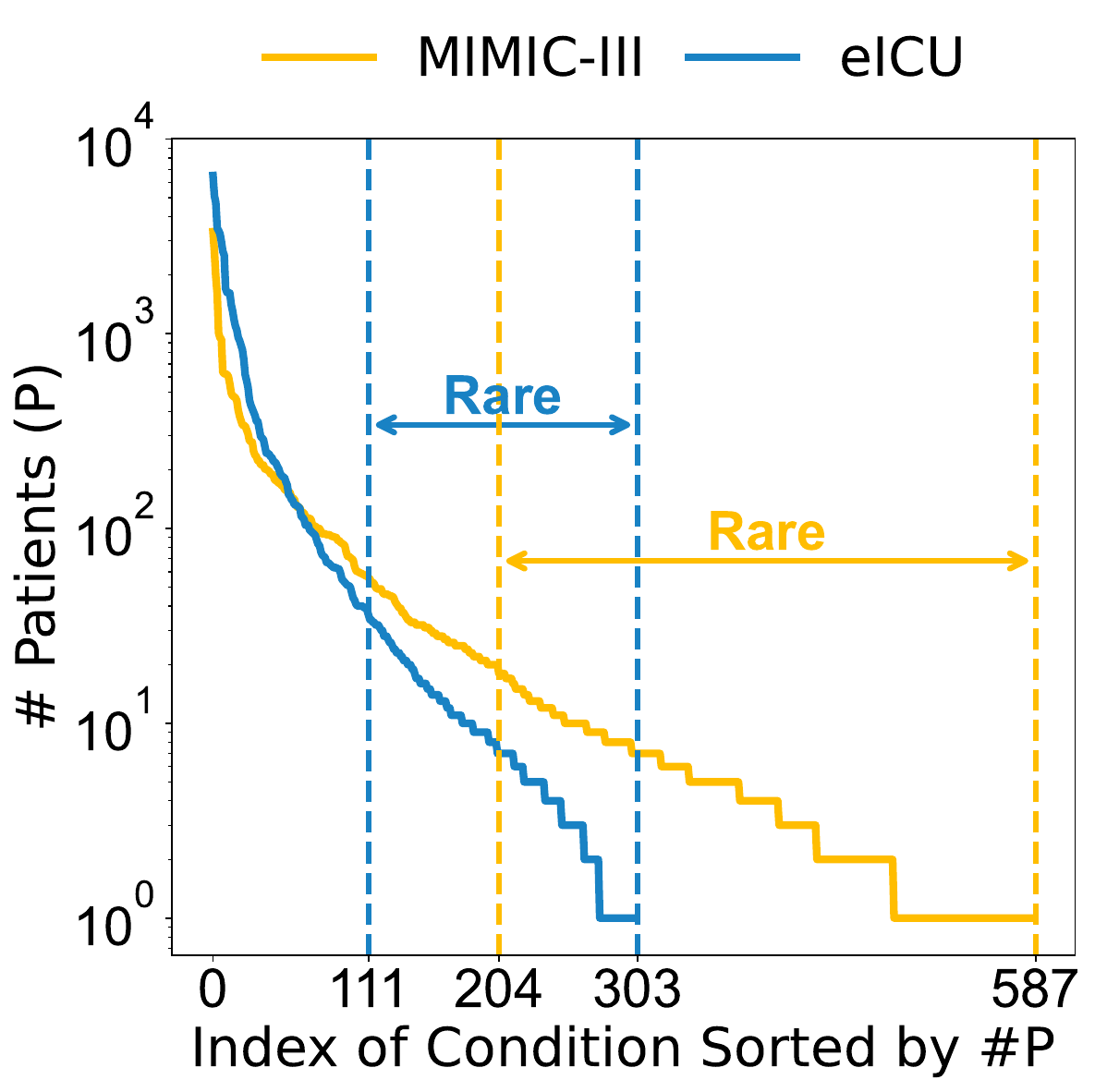}
    \end{subfigure}
    \hspace{0.6cm}
    \begin{subfigure}[b]
    {0.55\textwidth}
        \caption*{(b)}
        \vspace{-0.1cm}
        \centering
        \includegraphics[width=\textwidth]{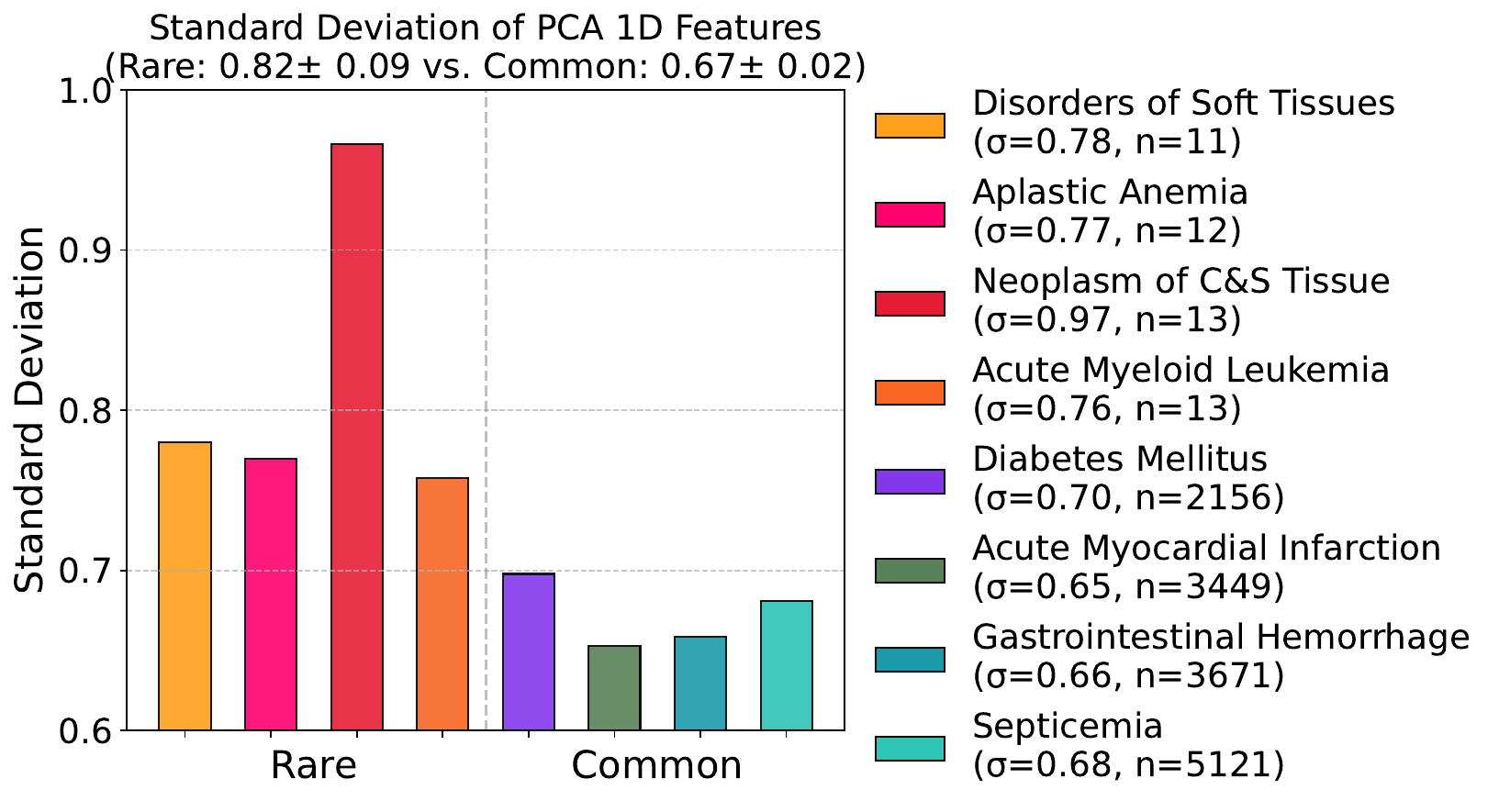}

    \end{subfigure}
    \vspace{-5px}
    \caption{\textbf{Data scarcity and intra-condition heterogeneity of rare conditions in ICU datasets.} (a) Condition prevalence in the MIMIC-III and eICU datasets. Conditions are indexed and labelled as \textit{Rare} if their prevalence is fewer than one case per 2,000 patients. In total, 383 of 587 conditions in MIMIC-III and 192 of 303 conditions in eICU are considered rare, highlighting the significant proportion of rare conditions in the ICU. (b) Histogram of standard deviation ($\sigma$) values of one-dimensional variables obtained by applying Principal Component Analysis (PCA) to patient variables for four common and four rare conditions in eICU. Here, $n$ denotes the number of patients for each condition. The mean $\sigma$ for rare conditions (0.82 $\pm$ 0.09) is higher and more variable compared to common conditions (0.67 $\pm$ 0.02), illustrating greater intra-condition heterogeneity among rare conditions.}
    \label{fig:challenges}
\end{figure}

We identify two challenges posed by rare conditions in the ICU for DL-based clinical prediction (Figure~\ref{fig:challenges}):

\begin{itemize}[leftmargin=10px]

    \item \textbf{Data Scarcity}: The low prevalence of rare conditions results in limited samples within the EHR. Figure~\ref{fig:challenges}(a) shows that 383 and 192 conditions in MIMIC-III\cite{johnson2016mimic} and eICU\cite{pollard2018eicu}, respectively, have a prevalence of less than 1 in 2,000 cases. The scarcity of data impedes the learning of robust patterns, often resulting in overfitting the model or inadequate generalisation \cite{aliferis2024overfitting, alzubaidi2023survey}.

    \item \textbf{intra-condition heterogeneity}: Rare conditions often display more heterogeneous measurement than common conditions\cite{phillips2024time}. Figure~\ref{fig:challenges}(b) Common conditions exhibit a more concentrated distribution, while rare conditions exhibit a notably wider distribution. The inherent diversity of rare conditions challenges the robustness of DL methods, especially with limited training samples \cite{liu2022natural, banerjee2023machine}.
\end{itemize}

\section{Data Processing}
\label{appendix sec:data processing}

\begin{table}[h!]
\centering
\footnotesize
\begin{threeparttable}
\caption{Extracted variables from MIMIC-III and eICU}
\label{table:variables}
\begin{tabular}{c|l c c l}
\hline
\textbf{Category}         & \textbf{variables} & \textbf{MIMIC-III} & \textbf{eICU} & \textbf{Description}  \\ \hline
\multirow{3}{*}{Demographics} & Age & $\checkmark$  &$\checkmark$  &The age of the patient at the time of hospital admission. \\
    & Gender & $\checkmark$ & $\checkmark$ &The gender of the patient (Male or Female). \\
    & Race & $\checkmark$ & $\checkmark$&The racial or ethnic category of the patient. \\
\hline
\multirow{7}{*}{Vital Sign} & Heart Rate   &$\checkmark$ & $\checkmark$ & Measures the rate of heartbeats per minute.               \\ 
                            & Systolic/Diastolic BP      &$\checkmark$ & $\checkmark$ & Measures systolic and diastolic blood pressure.           \\ 
                            & Arterial Pressure    &$\checkmark$ & $\checkmark$ & Calculates the average blood pressure during one cycle.   \\ 
                            & Respiratory Rate           &$\checkmark$ & $\checkmark$ &Measures the number of breaths per minute.                \\ 
                            & SpO2, Peripheral     &$\checkmark$ & $\checkmark$ & Measures blood oxygen saturation levels.                  \\ 
                            & Temperature &$\checkmark$ & $\checkmark$ &Measures body temperature in Celsius.   \\
                            & Bedside Glucose  &$\checkmark$ & $\checkmark$ &Monitors blood glucose levels.    \\ \hline
\multirow{19}{*}{Lab Test}  & Albumin    &$\checkmark$ &  $\checkmark$& A protein in blood plasma, used to assess liver and kidney function. \\ 
                            & Anion Gap   &$\checkmark$ &$\checkmark$ & A measure of ions in the blood, used to detect acid-base imbalances. \\ 
                            & Bands   &$\checkmark$ & & Immature white blood cells, indicative of infection or inflammation. \\ 
                            & Bicarbonate    &$\checkmark$ & $\checkmark$ & A buffer in blood, used to assess acid-base status. \\ 
                            & Bilirubin    &$\checkmark$ & $\checkmark$& A product of red blood cell breakdown. Used to evaluate liver function. \\ 
                            & Blood Urea Nitrogen   &$\checkmark$ &  & Indicates kidney function and hydration status. \\ 
                            & Chloride &$\checkmark$ &$\checkmark$  & An electrolyte, used to assess hydration and acid-base balance. \\
                            & Creatinine  &$\checkmark$ &$\checkmark$ & A waste product from muscle metabolism. \\ 
                            & Glucose   &$\checkmark$ & & Blood sugar level. \\ 
                            & Hematocrit &$\checkmark$ & $\checkmark$& The proportion of red blood cells in the blood. \\
                            & Haemoglobin &$\checkmark$ & $\checkmark$& A protein in red blood cells, carries oxygen throughout the body. \\ 
                            & International Normalised Ratio  &$\checkmark$ &  &Used to evaluate blood coagulation. \\ 
                            & Lactate  &$\checkmark$ &  $\checkmark$& Indicates tissue oxygenation and metabolic status. \\ 
                            & Platelet Count &$\checkmark$ &  $\checkmark$& Used to assess blood clotting ability. \\ 
                            & Potassium &$\checkmark$ &  & An electrolyte, critical for heart and muscle function. \\ 
                            & Prothrombin Time &$\checkmark$ &  $\checkmark$& Measures blood clotting speed, used to monitor anticoagulant therapy. \\ 
                            & Partial Thromboplastin Time  &$\checkmark$ &  $\checkmark$& Measures clotting, used to detect bleeding disorders. \\ 
                            & Sodium  &$\checkmark$ &  $\checkmark$& An electrolyte, critical for fluid balance and nerve function. \\
                            & White Blood Cell Count  &$\checkmark$ &  & Indicates immune response and infection. \\ \hline
\end{tabular}
\begin{tablenotes}
            \item[1] A checkmark ($\checkmark$) indicates that the variable is available and was extracted from the corresponding dataset.
\end{tablenotes}
\end{threeparttable}
\end{table}

We extracted demographic, vital sign, and laboratory test variables from MIMIC-III and eICU, as detailed in the Supplementary Table~\ref{table:variables}. The data extraction procedure involved several stages, including the selection of patient cohorts, the aggregation of time-series variables, the imputation of missing values, and the normalisation of the data.

First, we excluded patients with insufficient data availability, specifically, patients with hospital stays shorter than 48 hours in the MIMIC-III dataset and ICU stays shorter than 24 hours in the eICU dataset. Subsequently, we assigned a primary diagnosis to each patient based on the first three digits of their ICD-9-CM codes. Conditions with fewer than ten patients were excluded to ensure adequate sample sizes for robust evaluations. Following these criteria, the resulting dataset included 38,360 patient samples from MIMIC-III and 72,536 from eICU. Each dataset was then partitioned into training, validation, and test subsets using a 4:1:1 ratio, stratifying patients by condition to ensure representative distributions across all splits. After cohort selection, we extracted vital signs and laboratory test measurements from structured EHR databases. For the MIMIC-III dataset, we collected the available measurements within the last 48 hours after hospital admission. For the eICU dataset, the data were extracted from the initial 24 hours after admission to the ICU. Demographic information, including age, gender, and race, was obtained directly from patient metadata. To construct structured time-series representations, the extracted variables were aggregated into consistent temporal intervals, following established benchmark methodologies \cite{gupta2022extensive, harutyunyan2019multitask}. Specifically, MIMIC-III data were segmented into 24 two-hour intervals during the last 48 hours of admission, calculating the mean values within each interval. Similarly, eICU data were segmented into 24 one-hour intervals that cover the first 24 hours of admission to the ICU. Missing measurements within these intervals initially remained as missing values. To address missing data, we implemented widely accepted imputation techniques \cite{van2023yet, hyland2020early}. Forward imputation was applied for the first time, replacing missing values with the most recent available measurement from the same patient. If forward imputation was not possible due to a lack of previous data, backwards imputation was used, substituting the next available measurement. Any residual missing values after these two steps were filled in with the mean of the corresponding variable in the training set. This approach ensured the independence between the training set and the validation/test sets. Lastly, to standardise variable scales across datasets, continuous variables underwent Z-score normalisation using the mean and standard deviation computed exclusively from the training set. 

\begin{table*}[h]
    \centering
    \fontsize{7.5}{8.5}\selectfont
    \caption{Statistics for Selected Rare Conditions in MIMIC-III}
    \label{tab:mimic_rare_stats}
    \begin{tabular}{p{4.5cm}|cccccc}
        \hline
        Conditions (ICD-9-CM Code Level3) & \# Patients & \# Hospital Visits & Age & Female Rate & 90-Day Mortality & 30-Day Readmission Rate \\
        \hline
        Mycoses (117) & 13 & 13 & 50.17 & 0.46 & 0.23 & 0.08 \\
        Anemia (280) & 12 & 12 & 67.70 & 0.42 & 0.25 & 0.25 \\
        Hodgkin's Disease (201) & 12 & 12 & 55.44 & 0.58 & 0.08 & 0.08 \\
        Neoplasm of Digestive and Respiratory Systems (235) & 12 & 12 & 68.60 & 0.67 & 0.25 & 0.08 \\
        Bronchiectasis (494) & 11 & 11 & 72.49 & 0.55 & 0.18 & 0.09 \\
        Herpetic Whitlow (054) & 11 & 11 & 51.43 & 0.36 & 0.18 & 0.18 \\
        Malignant Neoplasm of Tongue (141) & 10 & 10 & 60.06 & 0.30 & 0.10 & 0.00 \\
        Neoplasms of Unspecified Nature (239) & 10 & 10 & 64.30 & 0.30 & 0.10 & 0.10 \\
        Effects of Reduced Temperature (991) & 10 & 10 & 59.17 & 0.30 & 0.20 & 0.10 \\
        Open Wound of Neck (874) & 10 & 10 & 43.38 & 0.30 & 0.10 & 0.10 \\
        \hline
    \end{tabular}
\end{table*}

\begin{table*}[h]
    \centering
    \fontsize{7.5}{8.5}\selectfont
    \caption{Statistics for Selected Rare Conditions in eICU}
    \label{tab:eicu_rare_stats}
    \begin{tabular}{p{5cm}|cccccc}
        \hline
        Condition (ICD-9-CM Code Level3) & \# Patients & \# ICU Stays & Age & Female Rate & ICU Mortality & Median ICU LoS (Q1-Q3, days) \\
        \hline
        Meningitis (322) & 14 & 16 & 64.88 & 0.57 & 0.06 &  2.9 (2.1, 5.4) \\
        Abscess of Lung and Mediastinum (513) & 14 & 14 & 51.29 & 0.36 & 0.07 & 3.8 (2.0, 5.5) \\
        Acute Myeloid Leukemia (205)  & 13 & 13 & 58.23 & 0.54 & 0.23 & 2.2 (1.6, 4.4) \\
        Acute Pericarditis (420)& 12 & 14 & 61.00 & 0.33 & 0.21 & 2.0 (1.3, 2.9) \\
        Malignant Neoplasm of Connective and Other Soft Tissue (171) & 11 & 13 & 63.08 & 0.18 & 0.23 & 2.9 (1.9, 3.7) \\
        Aplastic Anemia (284) & 11 & 12 & 61.75 & 0.64 & 0.08 & 1.9 (1.5, 3.1) \\
        Effects of Reduced Temperature (991) & 11 & 11 & 62.45 & 0.45 & 0.09 & 2.7 (1.6, 3.4) \\
        Other Disorders of Soft Tissues (729) & 10 & 11 & 62.09 & 0.70 & 0.09 & 2.0 (1.4, 2.7) \\
        Malignant Neoplasm of Gallbladder and Extrahepatic Bile Ducts (156) & 10 & 11 & 52.91 & 0.60 & 0.09 & 1.7 (1.3, 2.3) \\
        Fracture of Base of Skull (801) & 10 & 10 & 51.60 & 0.50 & 0.20 & 2.7 (1.5, 3.6) \\
        \hline
    \end{tabular}
\end{table*}

\section{Implementation Details}
\label{sec:implement}

\subsection{Model Architecture}
We developed the KnowRare framework and all model-agnostic baseline methods using the LSTM as the backbone for encoding time-series data. Specifically, the LSTM encoder consisted of a single-layer network with a hidden dimension of 128. In addition to time-series inputs, demographic variables (age, gender, and race) were processed through a two-layer Multilayer Perceptron (MLP) with LeakyReLU activation functions, also having a hidden dimension of 128. The resulting demographic embeddings were then concatenated with the final hidden state of the LSTM encoder, forming the latent representation. Subsequently, this combined representation was fed into a two-layer MLP classifier to predict clinical outcomes. Other modules of the KnowRare framework, including the decoder used during pre-training and the discriminator used for domain adaptation, were similarly implemented as two-layer MLP networks.

\subsection{Hyperparameters}
\begin{table}[h!]
    \centering
    \caption{Hyperparameter Search Space for Different Methods}
    \label{tab:sweep_configs}
    \begin{tabular}{l l l}
        \hline
        \textbf{Method} & \textbf{Hyperparameter} & \textbf{Search Space} \\
        \hline
        \multirow{2}{*}{LSTM} & Learning Rate & Uniform(1e-5, 1e-3) \\
                                  & Batch Size & \{16, 32, 64, 128\} \\
        \hline
        \multirow{2}{*}{Transformer} & Learning Rate & Uniform(1e-5, 1e-3) \\
                                  & Batch Size & \{16, 32, 64, 128\} \\
        \hline
        \multirow{2}{*}{RETAIN} & Learning Rate & Uniform(1e-5, 1e-3) \\
                                  & Batch Size & \{16, 32, 64, 128\} \\     
        \hline
        \multirow{2}{*}{MetaPred} & Meta Learning Rate  & Uniform(1e-5, 1e-3) \\
                                  & Inner Learning Rate  & Uniform(1e-5, 1e-3) \\
                                  & Batch Size & \{16, 32, 64, 128\} \\
                                  & Inner Loss Coefficient & \{0.1, 1.0, 10.0\} \\
        \hline
        \multirow{2}{*}{RareMed} & Learning Rate  & Uniform(1e-5, 1e-3) \\
                                  & Batch Size & \{16, 32, 64, 128\} \\
        \hline
        \multirow{2}{*}{SMART} & Learning Rate & Uniform(1e-5, 1e-3) \\
                                  & Batch Size & \{16, 32, 64, 128\} \\
        \hline
        \multirow{2}{*}{Stable-CRP} & Learning Rate & Uniform(1e-5, 1e-3) \\
                                    & Batch Size & \{16, 32, 64, 128\} \\
        \hline
        \multirow{5}{*}{FADA} & Learning Rate & Uniform(1e-5, 1e-3) \\
                              & Discriminator Learning Rate  & Uniform(1e-5, 1e-3) \\
                              & Batch Size & \{16, 32, 64, 128\} \\
                              & Adversarial Loss Coefficient & \{0.1, 0.5, 1.0, 2.0, 10.0\} \\
                              & Discriminator Update Frequency & Uniform(1, 10) \\
        \hline
        \multirow{5}{*}{AdvDiag} & Learning Rate & Uniform(1e-5, 1e-3) \\
                                 & Discriminator Learning Rate & Uniform(1e-5, 1e-3) \\
                                 & Batch Size & \{16, 32, 64, 128\} \\
                                 & Adversarial Loss Coefficient & \{0.1, 0.5, 1.0, 2.0, 10.0\} \\
                                 & Discriminator Update Frequency & Uniform(1, 10) \\
        \hline
        \multirow{2}{*}{MANYDG} & Learning Rate & Uniform(1e-5, 1e-3) \\
                                  & Batch Size & \{16, 32, 64, 128\} \\
        \hline
        \multirow{5}{*}{KnowRare} & Learning Rate & Uniform(1e-5, 1e-3) \\
                               & Discriminator Learning Rate & Uniform(1e-5, 1e-3) \\
                               & Batch Size & \{16, 32, 64, 128\} \\
                               & Adversarial Loss Coefficient & \{0.005, 0.01, 0.02, 0.1\} \\
                               & Discriminator Update Frequency & Uniform(1, 5)\\
        \hline
    \end{tabular}
\end{table}

The hyperparameters of all baseline methods are determined using Bayesian optimisation with 30 iterations based on their performance on the validation set. Baseline models that include a pre-training stage (FADA, AdvDiag, SMART, RareMed, and KnowRare) are pre-trained on the entire training set. For KnowRare, we select 10\% of all conditions for joint adversarial domain adaptation. All methods are trained for a maximum of 100 epochs, with early stopping applied if no performance improvement is observed for 10 consecutive epochs. We repeated our experiment five times with different random seeds, reporting the mean and standard deviation (std) for each evaluation metric. The hyperparameters for KnowRare and each baseline method are detailed in Table~\ref{tab:sweep_configs}. We utilised the Weights \& Biases platform to perform an extensive hyperparameter search. The optimal hyperparameters were selected based on the highest AUPRC achieved in the validation set.

\subsection{Training Settings}
We employ the Adam optimiser for training, initialising with a warm-up phase of a fixed learning rate of 10 epochs. Following this, we apply an exponential decay to the learning rate at a rate of 0.95 after each subsequent epoch. To prevent overfitting and ensure efficient training, we implement an early stopping mechanism that stops training if there is no improvement in validation performance over 10 consecutive epochs. The optimiser is configured with $\beta_1 = 0.9$, $\beta_2 = 0.999$, and $\epsilon = 10^{-8}$, adhering to the default parameters settings. 

\section{Results of knowledge-guided domain selection}

Table~\ref{tab:source_conditions} presents the selected source conditions corresponding to the target rare conditions, as identified through the proposed knowledge-guided domain selection method.

\clearpage
\begin{center}
\small
\tablecaption{The selected source conditions for each rare condition on MIMIC-III and eICU}
\label{tab:source_conditions}
\tablefirsthead{
  \hline
  \textbf{Rare Condition (ICD-9-CM Code Level 3)} & \textbf{Dataset} & \textbf{Source Conditions (ICD-9-CM Code Level 3)} \\
  \hline
}
\tablehead{
  \hline
  \textbf{Rare Condition (ICD-9-CM Code Level 3)} & \textbf{Dataset} & \textbf{Source Conditions (ICD-9-CM Code Level 3)} \\
  \hline
}

\tabletail{
    \hline
    \multicolumn{3}{r}{\small\sl continued on next page} \\
}

\tablelasttail{
    \hline
    \multicolumn{3}{r}{\small\sl end of the table} \\
}
\begin{supertabular}{p{3cm} c p{12.5cm}}
Mycoses (117) & MIMIC-III & Chronic airway obstruction (510), Other diseases of lung (724), Other diseases of blood and blood-forming organs (289), Other diseases of arteries, arterioles, and capillaries (456), Other disorders of circulatory system (447), Congenital anomalies of heart (745), Other diseases of bone and cartilage (721), Malignant neoplasm of other and unspecified sites (196), Chronic liver disease and cirrhosis (570), Migraine (346), Gastritis and duodenitis (535), Other diseases of blood and blood-forming organs (572), Other diseases of lung (420), Other disorders of circulatory system (437), Other diseases of blood and blood-forming organs (292), Congenital anomalies of heart (746), Other diseases of lung (873), Other diseases of veins and lymphatics, and other diseases of circulatory system (455), Other diseases of lung (710), Other diseases of blood and blood-forming organs (202), Other disorders of circulatory system (444), Other diseases of blood and blood-forming organs (255), Other diseases of blood and blood-forming organs (324), Other diseases of genitourinary system (593), Chronic airway obstruction (494), Gastric ulcer (430), Gastric ulcer (531) \\
\hline
Iron Deficiency Anemia (280) & MIMIC-III & Other diseases of blood and blood-forming organs (507), Epilepsy (345), Other diseases of nervous system (349), Other diseases of respiratory system (519), Chronic ulcer of skin (707), Other diseases of genitourinary system (511), Chronic liver disease and cirrhosis (570), Intestinal infection due to other organisms (008), Conduction disorders (426), Other diseases of blood and blood-forming organs (572), Acute myocardial infarction (410), Alcohol dependence syndrome (303), Other diseases of circulatory system (459), Pneumonia, organism unspecified (482), Other disorders of intestine (569), Symptoms involving respiratory system and other chest symptoms (786), Other diseases of veins and lymphatics, and other diseases of circulatory system (453), Endocarditis (424), Other forms of heart disease (440), Cardiomyopathy (425), Other diseases of blood and blood-forming organs (288), Other diseases of genitourinary system (593), Osteoarthrosis and allied disorders (715), Gastrointestinal hemorrhage (578), Other disorders of bone and cartilage (733), Other disorders of circulatory system (443), Other forms of heart disease (411) \\
\hline
Hodgkin's Disease (201) & MIMIC-III &Gastritis and duodenitis (534), Meningitis due to other organisms (047), Other diseases of bone and cartilage (803), Spina bifida (742), Poisoning by antibiotics (980), Injury to blood vessels of thorax (901), Other diseases of blood and blood-forming organs (862), Pneumonia, organism unspecified (480), Other diseases of synovium, tendon, and bursa (711), Acute pharyngitis (464), Other unspecified infectious and parasitic diseases (991), Poisoning by other central nervous system depressants and anesthetics (972), Other diseases of digestive system (212), Intestinal infection due to other organisms (009), Other diseases of heart (404), Diseases of mitral and aortic valves (394), Other unspecified infectious and parasitic diseases (994), Open wound of other and unspecified sites, except limbs (874), Hypertensive heart disease (402), Carcinoma in situ of breast and genitourinary system (233), Inguinal hernia (550), Anomalies of peripheral vascular system (747), Meningitis due to other organisms (136), Poisoning by agents primarily affecting blood constituents (966), Malignant neoplasm of other and unspecified sites (183), Malignant neoplasm of other and unspecified sites (164), Diseases of mitral and aortic valves (395) \\
\hline
Neoplasm of Digestive and Respiratory Systems (235) & MIMIC-III & Other diseases of male genital organs (608), Other diseases of bone and cartilage (803), Meningitis due to other organisms (047), Malignant neoplasm of other and unspecified sites (171), Open wound of other and unspecified sites, except limbs (881), Malignant neoplasm of other and unspecified sites (158), Malignant neoplasm of other and unspecified sites (161), Other diseases of blood and blood-forming organs (862), Pneumonia, organism unspecified (480), Malignant neoplasm of other and unspecified sites (152), Acute pharyngitis (464), Spina bifida (738), Poisoning by other central nervous system stimulants (982), Injury to blood vessels of abdomen (864), Other diseases of heart (404), Intestinal infection due to other organisms (009), Other diseases of blood and blood-forming organs (866), Other diseases of respiratory system (513), Injury to blood vessels of abdomen (902), Other diseases of blood and blood-forming organs (200), Open wound of other and unspecified sites, except limbs (874), Other diseases of blood and blood-forming organs (228), Other diseases of blood and blood-forming organs (209), Other diseases of blood and blood-forming organs (674), Malignant neoplasm of lip,, oral cavity, and pharynx (141), Malignant neoplasm of other and unspecified sites (183), Injury to blood vessels of head and neck (854) \\ \hline

Bronchiectasis (494) & MIMIC-III & Other diseases of male genital organs (608), Rheumatic mitral valve disease (398), Pneumonia due to other specified bacteria (481), Other disorders of circulatory system (447), Malignant neoplasm of pancreas (157), Other alveolar and parietoalveolar pneumonopathies (516), Infection of kidney (590), Fracture of neck of femur (820), Fracture of radius and ulna (813), Fracture of lumbar spine and pelvis (806), Lymphoid leukemia (204), Pneumonia, organism unspecified (480), Osteoarthrosis and allied disorders (711), Multiple myeloma and immunoproliferative neoplasms (203), Malignant neoplasm of kidney and other unspecified urinary organs (189), Congenital anomalies of heart (746), Benign neoplasm of uterus (225), Systemic lupus erythematosus (710), Abscess of lung and mediastinum (513), Gastric ulcer (432), Other diseases of male genital organs (235), Gastric ulcer (430), Anomalies of peripheral vascular system (747), Polyarteritis nodosa and allied conditions (446), Fracture of base of skull (801), Disorders of lipoid metabolism (277), Transient cerebral ischemia (435) \\ \hline

Herpetic Whitlow (054) & MIMIC-III & Empyema (510), Other disorders of soft tissue (728), Osteomyelitis, periostitis, and other infections involving bone (730), Other diseases of arteries, arterioles, and capillaries (456), Congenital anomalies of heart (745), Chronic bronchitis (491), Disorders of adrenal glands (253), Regional enteritis (555), Internal injury of chest, abdomen, and pelvis (860), Chronic liver disease and cirrhosis (570), Migraine (346), Gastritis and duodenitis (535), Acute vascular insufficiency of intestine (557), Acute pericarditis (420), Toxic effect of other substances, chiefly nonmedicinal as to source (292), Internal injury of chest, abdomen, and pelvis (861), Ill-defined descriptions and complications of heart disease (429), Hemorrhoids (455), Other disorders of arteries, arterioles, and capillaries (442), Other diseases of blood and blood-forming organs (202), Epistaxis (784), Other disorders of circulatory system (444), Disorders of adrenal glands (255), Gastric ulcer (430), Poisoning by analgesics, antipyretics, and antirheumatics (965), Neoplasms of uncertain behavior (238), Diseases of pericardium (397) \\ \hline

Malignant neoplasm of Tongue (141) & MIMIC-III & Gastritis and duodenitis (534), Other specified diseases due to viruses (088), Spina bifida (742), Open wound of other and unspecified sites, except limbs (881), Poisoning by antibiotics (980), Malignant neoplasm of other and unspecified sites (158), Malignant neoplasm of other and unspecified sites (161), Other diseases of blood and blood-forming organs (862), Pneumonia, organism unspecified (480), Malignant neoplasm of other and unspecified sites (152), Acute pharyngitis (464), Spina bifida (738), Poisoning by other central nervous system stimulants (982), Poisoning by other central nervous system depressants and anesthetics (972), Poisoning by agents primarily affecting blood constituents (962), Other diseases of heart (404), Other diseases of blood and blood-forming organs (866), Other diseases of blood and blood-forming organs (200), Open wound of other and unspecified sites, except limbs (874), Trigeminal nerve disorders (350), Other diseases of blood and blood-forming organs (209), Other diseases of blood and blood-forming organs (228), Other diseases of male genital organs (235), Other diseases of blood and blood-forming organs (674), Malignant neoplasm of other and unspecified sites (164), Malignant neoplasm of other and unspecified sites (183), Injury to blood vessels of head and neck (854) \\ \hline

Neoplasms of Unspecified Nature (239) & MIMIC-III & Gastritis and duodenitis (534), Meningitis due to other organisms (047), Anomalies of digestive system (751), Injury to blood vessels of thorax (901), Malignant neoplasm of brain (191), Injury to blood vessels of head and neck (851), Malignant neoplasm of liver and intrahepatic bile ducts (156), Other diseases of blood and blood-forming organs (862), Pneumonia, organism unspecified (480), Acute pharyngitis (464), Malignant neoplasm of kidney and other unspecified urinary organs (189), Poisoning by other central nervous system stimulants (982), Poisoning by other central nervous system depressants and anesthetics (972), Neoplasms of uncertain behavior (237), Other diseases of digestive system (212), Malignant neoplasm of other and unspecified sites (193), Other diseases of heart (404), Injury to blood vessels of abdomen (902), Other unspecified infectious and parasitic diseases (994), Encephalitis, myelitis, and encephalomyelitis (323), Hypertensive heart disease (402), Carcinoma in situ of breast and genitourinary system (233), Spinal cord injury without evidence of spinal bone injury (952), Poisoning by other agents primarily affecting blood constituents (970), Anomalies of peripheral vascular system (747), Meningitis due to other organisms (136), Polyarteritis nodosa and allied conditions (446) \\ \hline

Effects of Reduced Temperature (991) & MIMIC-III & Meningitis due to other organisms (047), Other diseases of bone and cartilage (803), Spina bifida (742), Open wound of other and unspecified sites, except limbs (881), Poisoning by antibiotics (980), Fracture of ankle (825), Malignant neoplasm of breast (174), Anomalies of digestive system (751), Injury to blood vessels of thorax (901), Malignant neoplasm of brain (191), Other diseases of blood and blood-forming organs (862), Poisoning by other central nervous system stimulants (982), Other diseases of blood and blood-forming organs (201), Other diseases of digestive system (212), Poisoning by agents primarily affecting blood constituents (962), Intestinal infection due to other organisms (009), Other diseases of heart (404), Diseases of mitral and aortic valves (394), Coagulation defects (666), Hypertensive heart disease (402), Trigeminal nerve disorders (350), Other diseases of blood and blood-forming organs (228), Acute appendicitis (540), Anomalies of peripheral vascular system (747), Meningitis due to other organisms (136), Malignant neoplasm of lip, oral cavity, and pharynx (141), Malignant neoplasm of other and unspecified sites (183) \\ \hline

Open Wound of Neck (874) & MIMIC-III & Other diseases of bone and cartilage (803), Other inflammatory conditions of skin (088), Spina bifida (742), Open wound of other and unspecified sites, except limbs (881), Poisoning by antibiotics (980), Malignant neoplasm of other and unspecified sites (161), Other diseases of blood and blood-forming organs (862), Pneumonia, organism unspecified (480), Malignant neoplasm of other and unspecified sites (152), Acute pharyngitis (464), Poisoning by other central nervous system stimulants (982), Poisoning by other central nervous system depressants and anesthetics (972), Poisoning by agents primarily affecting the cardiovascular system (962), Other diseases of heart (404), Intestinal infection due to other organisms (009), Other diseases of blood and blood-forming organs (866), Other unspecified infectious and parasitic diseases (994), Diseases of mitral and aortic valves (394), Delivery in a completely normal case (666), Trigeminal nerve disorders (350), Other diseases of blood and blood-forming organs (209), Other diseases of blood and blood-forming organs (228), Other diseases of blood and blood-forming organs (674), Malignant neoplasm of lip, oral cavity, and pharynx (141), Malignant neoplasm of other and unspecified sites (164), Malignant neoplasm of other and unspecified sites (183), Injury to blood vessels of head and neck (854) \\
\hline

Fracture of Base of Skull (801) & eICU & Empyema (510), Other disorders of soft tissue (728), Poisoning by other specified drugs and medicinal substances (968), Pancreatic disorders (576), Malignant neoplasm of rectum, rectosigmoid junction, and anus (154), Diseases of hard tissues of teeth (523), Other vascular insufficiencies of intestine (557), Toxic effect of other substances, chiefly nonmedicinal as to source (292), Fracture of shaft of femur (821), Secondary malignant neoplasm of respiratory and digestive systems (197), Other disorders of the urethra and urinary tract (599), Iron deficiency anaemias (280), Other hypertensive heart diseases (404), Intracerebral haemorrhage (431), Paralytic ileus (560), Injury to intra-abdominal organs (866), Other diseases of the  upper respiratory tract (478), Hereditary hemolytic anaemias (282), Polyarteritis nodosa and allied conditions (446) \\
\hline
Other Disorders of Soft Tissues (729) & eICU & Empyema (510), Aplastic anaemia (284), Poisoning by antibiotics (980), Pancreatic disorders (576), Cholelithiasis (574), Malignant neoplasm of female breast (174), Diseases of hard tissues of teeth (523), Infections of the kidney (590), Toxic effect of other substances, chiefly nonmedicinal as to source (292), Other fracture of skull (853), Malignant neoplasm of other and unspecified sites (193), Cholecystitis (575), Open wound of other and unspecified sites, except limbs (874), Other diseases of the upper respiratory tract (478), Complications affecting specified body systems (997), Fracture of vault of the skull (800), Diseases of the pancreas (577), Poisoning by analgesics, antipyretics, and antirheumatics (965) Thyroid gland disorders(242) \\
\hline
Malignant Neoplasm of Gallbladder and Extrahepatic Bile Ducts (156) & eICU & Esophageal varices (456), Pancreatic disorders (576), Cholelithiasis (574), Other disorders of nervous system (349), Malignant neoplasm of female breast (174), Fracture of vertebral column without mention of spinal cord injury (805), Other diseases of lung (516), Malignant neoplasm of other and unspecified sites (161), Chronic liver disease and cirrhosis (570), Acute pericarditis (420), Chronic pulmonary heart disease (416), Inflammatory and toxic neuropathy (357), Bulbus cordis anomalies and anomalies of cardiac septal closure (746), Peritonitis (567), Other hypertensive heart disease (404), Other disorders of stomach and duodenum (537), Other disorders of adrenal glands (255), Asthma (493) \\
\hline
Malignant Neoplasm of Connective and Other Soft Tissue(171) & eICU & Osteomyelitis, periostitis, and other infections involving bone (730), Diverticula of the intestine (562), Pancreatic disorders (576), Malignant neoplasm of rectum, rectosigmoid junction, and anus (154), Malignant neoplasm of pancreas (157), Other vascular insufficiencies of the intestine (557), Chronic pulmonary heart disease (416), Acute pharyngitis (464), Iron deficiency anaemias (280), Chronic liver disease and cirrhosis (571), Empyema and pneumothorax (510), Paralytic ileus (560), Symptoms involving respiratory system and other chest symptoms (786), Other diseases of the endocardium (424), Acute bronchitis and bronchiolitis (466), Other diseases of the upper respiratory tract (478), Diverticulosis and diverticulitis of colon (562), Other peripheral vascular diseases (443), Other forms of chronic ischemic heart disease (411) \\
\hline
Aplastic Anaemia (284) & eICU & Empyema (510), Osteomyelitis, periostitis, and other infections involving bone (730), Malignant neoplasm of the bladder (188), Diseases of hard tissues of teeth (523), Other ill-defined cerebrovascular disease (437), Pyogenic arthritis (711), Other disorders of soft tissue (729), Occlusion and stenosis of precerebral arteries (433), Acute ill-defined, cerebrovascular disease (436), Other complications of procedures (998), Open wound of other and unspecified sites, except limbs (874), Other diseases of upper respiratory tract (478), Burns classified according to extent of body surface involved (946), Hereditary hemolytic anaemias (282), Diseases of the pancreas (577), Poisoning by analgesics, antipyretics, and antirheumatics (965), Polyarteritis nodosa and allied conditions (446), Thyroid gland disorders (242), Other forms of chronic ischemic heart disease (411) \\
\hline
Effects of reduced temperature (991) & eICU & Empyema (510), Pneumonia, organism unspecified (481), Aplastic anaemia (284), Poisoning by antibiotics (980), Diseases of hard tissues of teeth (523), Malignant neoplasm of other and unspecified sites (161), Acute myocardial infarction (410), Other ill-defined cerebrovascular disease (437), Fracture of shaft of femur (821), Other diseases of the peripheral nervous system (358), Fracture of face bones (802), Generalised ischemic heart disease (414), Complications peculiar to certain specified procedures (996), Acute ill-defined, cerebrovascular disease (436), Complications affecting specified body systems (997), Diseases of the pancreas (577), Thyroid gland disorders (242), Acute bronchitis and bronchiolitis (466), Meningitis due to other organisms (322) \\
\hline
Acute Pericarditis (420) & eICU & Pneumonia, organism unspecified (481), Osteomyelitis, periostitis, and other infections involving bone (730), Esophageal varices (456), Acute myeloid leukaemia (205), Pancreatic disorders (576), Malignant neoplasm of female breast (174), Other unspecified infectious and parasitic diseases (999), Toxic effect of other substances, chiefly nonmedicinal as to source (292), Pyogenic arthritis (711), Chronic pulmonary heart disease (416), Pneumonia due to other specified bacteria (482), Other diseases of the peripheral nervous system (358), Fracture of face bones (802), Paralytic ileus (560), Injury to intra-abdominal organs (866), Other diseases of the upper respiratory tract (478), Other arterial embolism and thrombosis (444), Asthma (493), Malignant neoplasm of the stomach (151) \\
\hline
Acute Myeloid Leukemia (205) & eICU & Diseases of mitral and aortic valves (394), Other diseases of the central nervous system (344), Fracture of radius and ulna (813), Conduction disorders (426), Acute pericarditis (420), Chronic pulmonary heart disease (416), Fracture of femur (821), Drug-induced mental disorders (292), Pneumonia, organism unspecified (482), Injury to blood vessels of the abdomen (864), Fracture of face bones (802), Other symptoms involving head and neck (478), Hypertensive heart disease (402), Diseases of white blood cells (288), Other diseases of the blood and blood-forming organs (282), Gastric ulcer (531), Other disorders of the circulatory system (443), Thyroid gland disorders (242), Acute bronchitis and bronchiolitis (466) \\
\hline
Meningitis (322) & eICU & Empyema and pneumothorax (510), Pneumonia, organism unspecified (481), Other diseases of intestines and peritoneum (562), Malignant neoplasm of the bladder (188), Other inflammatory conditions of the skin (682), Malignant neoplasm of other and unspecified sites (161), Abscess of lung and mediastinum (572), Pneumonia, organism unspecified (480), Other disorders of the circulatory system (437), Other unspecified infectious and parasitic diseases (991), Cystitis (595), Pneumonia, organism unspecified (482), Appendicitis (541), Fracture of face bones (802), Other diseases of lung (512), Other diseases of stomach and duodenum (537), Other diseases of pancreas (577), Other disorders of urethra and urinary tract (599), Duodenal ulcer (533) \\
\hline
Abscess of Lung and Mediastinum (513) & eICU & Empyema and pneumothorax (510), Malignant neoplasm of nasal cavities, middle ear, and accessory sinuses (192), Other diseases of blood and blood-forming organs (253), Other diseases of respiratory system (519), Malignant neoplasm of other and unspecified sites (161), Fracture of radius and ulna (813), Complications peculiar to certain specified procedures (999), Coagulation defects (286), Alcoholic psychoses (291), Acute pericarditis (420), Septicemia (711), Renal failure (586), Injury to blood vessels of abdomen (866), Complications affecting specified body systems (996), Other diseases of muscle, ligament, and fascia (359), Other disorders of circulatory system (444), Other diseases of pancreas (577), Poisoning by analgesics, antipyretics, and antirheumatics (965), Disorders of thyroid gland (242) \\
\end{supertabular}
\end{center}

\label{SupEnd}

\end{document}